\documentclass[10pt,journal,compsoc]{IEEEtran}
\ifCLASSOPTIONcompsoc
  \usepackage[nocompress]{cite}
\else
  \usepackage{cite}
\fi
\ifCLASSINFOpdf
\else
\fi
\usepackage{booktabs} % For formal tables
\usepackage{subfig}
\usepackage{graphicx}
\usepackage{algorithm}
\usepackage{algpseudocode}% 
\usepackage{xcolor}
\usepackage{amsmath, amsthm}
\usepackage{amssymb}
\newcommand{\etal}{\textit{et al}. }

\usepackage{mathtools}
\usepackage{multirow}
\usepackage{graphics}
\usepackage{enumitem}
\usepackage{CJKutf8}
\usepackage{array}
\usepackage{diagbox}
\usepackage{svg}
\usepackage{url}
\usepackage{threeparttable}
\usepackage{comment}
\usepackage{xcolor,cite,etoolbox}
\definecolor{mypurple}{rgb}{0.4392, 0.1882, 0.6275}
\definecolor{darkblue}{rgb}{0.0, 0.0, 0.55}
\definecolor{darkgray}{rgb}{0.66, 0.66, 0.66}
\makeatletter 
\pretocmd\@bibitem{\color{black}\csname keycolor#1\endcsname}{}{\fail}
\newcommand \citecolor[1]{\@namedef{keycolor#1}{\color{red}}}
\makeatother
\begin{document}

% PL-DCP: A Pairwise Learning framework with Domain and Class Prototypes for EEG Emotion Recognition under Unseen Target Conditions
% MAT:A Multi-domain Aggregation Transfer Learning Framework for EEG Emotion Recognition with Domain-Class Prototype under Unseen Targets
\title{\textcolor{black}{Learning Domain- and Class-Disentangled Prototypes for Domain-Generalized EEG Emotion Recognition}}

\author{\IEEEauthorblockN{
% Weishan Ye\textsuperscript{1},
% Zhiguo Zhang\textsuperscript{1},
% Fei Teng,
% Min Zhang,
% Jianhong Wang,\\
% Dong Ni, 
% Fali Li,
% Peng Xu\textsuperscript{*}, and
% Zhen Liang\textsuperscript{*}}\\
Guangli Li,
Canbiao Wu,
Zhehao Zhou,
Na Tian,
Pengwen Dai,
Yuhao Wang,
Li Zhang\textsuperscript{*}, and
Zhen Liang\textsuperscript{*}}\\
\medskip
% <-this % stops a space

\IEEEcompsocitemizethanks{
\IEEEcompsocthanksitem Guangli Li, Canbiao Wu, Zhehao Zhou, and Na Tian are with the School of Biological Science and Medical Engineering, Hunan University of Technology, Zhuzhou 412008, China. E-mail: guangli010@hut.edu.cn, wucanbiao@m.scnu.edu.cn, 463805331@qq.com and tiann7@mail2.sysu.edu.cn.
\\
\IEEEcompsocthanksitem Pengwen Dai is with School of Cyber Science and Technology, Sun Yat-sen University, Guangzhou, 510275, China. E-mail: daipw@mail.sysu.edu.cn.
\\
\IEEEcompsocthanksitem Yuhao Wang is with Center for Brain-Computer Interface and Brain-Inspired Intelligence, Beijing Institute of Technology, Beijing, 100081, China. E-mail: wangyuhao@bit.edu.cn.
\\
\IEEEcompsocthanksitem Li Zhang and Zhen Liang are with the School of Biomedical Engineering, Health Science Center, Shenzhen University, Shenzhen 518060, China, and also with the Guangdong Provincial Key Laboratory of Biomedical Measurements and Ultrasound Imaging, Shenzhen 518060, China, and Zhen Liang also with the Shenzhen Pengrui Brain Science Technology, Shenzhen  518060, China. E-mail: \{lzhang, janezliang\}@szu.edu.cn.
\\
\IEEEcompsocthanksitem  \textsuperscript{*}Corresponding author: Li Zhang and Zhen Liang.}}
\IEEEtitleabstractindextext{
\begin{abstract}
\textcolor{black}{Electroencephalography (EEG)-based emotion recognition plays a critical role in affective Brain–Computer Interfaces (aBCIs), yet its practical deployment remains limited by inter-subject variability, reliance on target-domain data, and unavoidable label noise. To address these challenges, we propose a Multi-domain Aggregation Transfer learning framework with domain–class prototypes (MAT) for emotion recognition under completely unseen target domains. MAT introduces a feature decoupling module to disentangle class-invariant domain features from domain-invariant class features, enabling more robust and interpretable EEG representations. A Hierarchical-Domain Aggregation (HDA) mechanism based on Maximum Mean Discrepancy (MMD) constructs superdomains to model shared distributional structures across subjects, while adaptive prototype updating refines domain and class prototypes to capture stable intrinsic representations. Moreover, a pairwise learning strategy reformulates classification as similarity estimation between sample pairs, effectively mitigating the effect of label noise. We conduct extensive experiments on four public EEG emotion datasets (SEED, SEED-IV, SEED-V, and SEED-VII). \textcolor{black}{The results show that the accuracy of MAT is improved by 2.78\%, 3.84\%, 4.45\%, and 2.21\% compared with the state-of-the-art (SOTA) unseen-target models.} Furthermore, the proposed framework generalizes effectively to clinical depression identification scenarios, validating its application potential in real-world settings. The source code is available at \textit{https://github.com/WuCB-BCI/MAT}}.
\end{abstract}

}

\maketitle

%\footnotetext[1]{\hspace{1mm}Equal contributions.}
\IEEEdisplaynontitleabstractindextext

% \fi
\IEEEpeerreviewmaketitle

\IEEEraisesectionheading{
\section{Introduction}
\label{sec:introduction}}
\IEEEPARstart{E}{motion} 
\textcolor{black}{is a complex physiological and psychological state arising from internal processes or external stimuli \cite{Duan_Zhu_Lu_2013}. It profoundly influences human cognition, behavior, and health \cite{2002Affective}. Accurately describing and recognizing emotional states has therefore become a core challenge in affective computing, a rapidly evolving interdisciplinary field integrating neuroscience, psychology, and artificial intelligence. Traditional emotion recognition has largely relied on external cues such as speech \cite{Tian_Moore_Lai_2017}, facial expressions \cite{Sun_Li_2016}, and body gestures \cite{Zhao_Wang_2013}, or internal physiological signals such as electrocardiography (ECG) \cite{Agrafioti_2012} and electroencephalography (EEG) \cite{ye2024semi}. Compared with external modalities, EEG signals provide intrinsic, real-time, and objective indicators of affective states, making them highly promising for affective Brain–Computer Interfaces (aBCIs) \cite{Ye2023AdaptiveSA}.} 

\begin{figure}[t]
\centering
\subfloat{\includegraphics[width=0.49\textwidth]{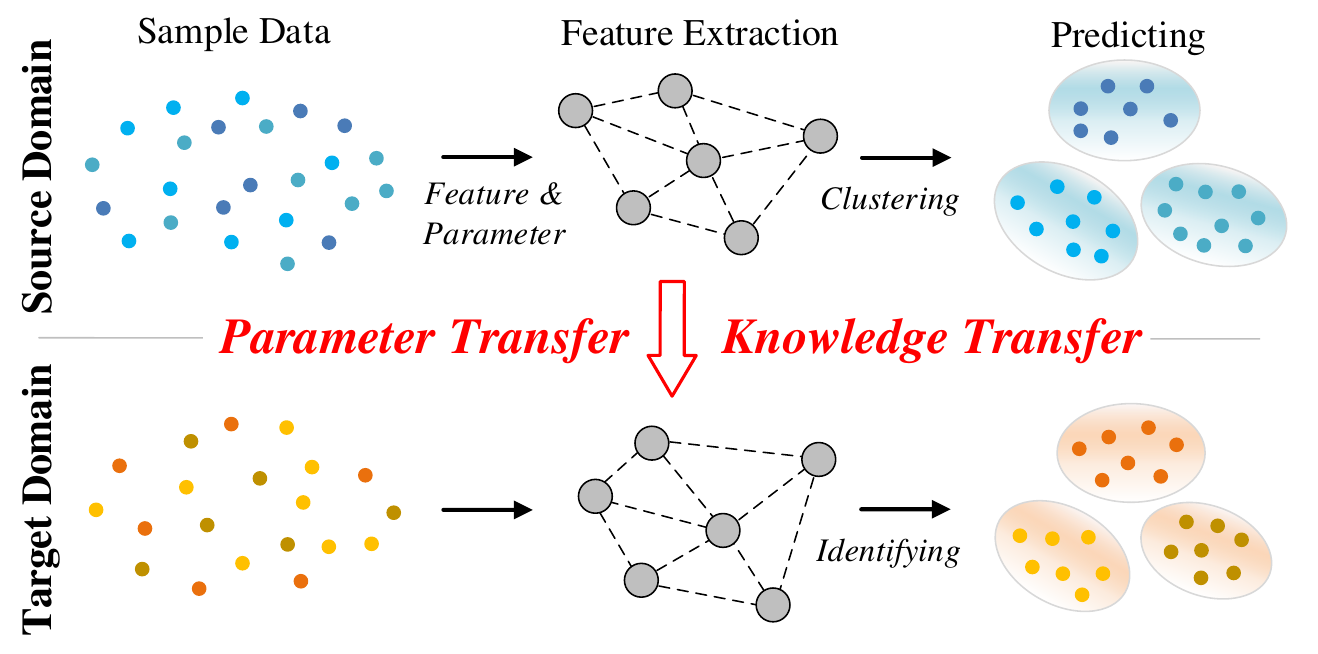}}
\caption{\textcolor{black}{Illustration of the transfer learning paradigm. Knowledge extracted from one or multiple source domains is transferred to a target domain to enhance feature alignment and model generalization across subjects.}}%迁移学习策略的基本框架，
\label{fig:Transfer_Learning}
\end{figure}

\textcolor{black}{Recent years have witnessed the growing success of deep transfer learning methods in EEG-based emotion recognition \cite{Li_Huan_2022}. As illustrated in Fig.~\ref{fig:Transfer_Learning}, transfer learning leverages labeled source domains and unlabeled target domains to transfer knowledge across distributions. Despite their effectiveness, existing models face three major challenges. \textbf{(1) Dependence on target-domain data.} Most transfer learning paradigms assume model parameter transferability across domains and require simultaneous access to both source and target data during training. This assumption severely limits generalization to unseen subjects, as models must be retrained for new individuals and often overfit to target-domain preferences \cite{zhou2024eegmatch,luo2024m3d}. \textbf{(2) Pronounced individual variability.} EEG signals exhibit strong subject dependency in emotional perception and neural expression \cite{paranjape2001eeg}. Such inter-subject heterogeneity complicates emotion modeling, necessitating approaches capable of learning domain-invariant yet class-discriminative representations that generalize across individuals \cite{Ye2023AdaptiveSA, liang2021eegfusenet}. \textbf{(3) Label noise in emotion induction.} Most EEG emotion datasets are elicited by videos, yet individual subjective responses to the same stimulus vary considerably, leading to weak correlations between physiological arousal and self-reported labels. Consequently, inevitable annotation noise degrades supervised models. Unlike traditional pointwise classification, pairwise learning formulations \cite{zhou2023pr,li2025unsupervised} exploit inter-sample relationships and exhibit superior robustness to noisy labels.} 
%To jointly address these challenges,

\textcolor{black}{\textcolor{black}{
Inspired by prior studies \cite{Zhang_2022_CVPR}, we hypothesize that EEG signals involve two complementary feature types based on physiological and cognitive evidence: domain-specific and class-specific representations. The former represents the characteristic shift caused by individual anatomical differences, while the latter represents the emotional commonality cross subjects. Thus, we propose a \textbf{M}ulti-domain \textbf{A}ggregation \textbf{T}ransfer learning framework with domain-class prototypes (MAT). Specifically, a feature decoupling module learns a domain-class dual-prototype representation, while a hierarchical domain aggregation mechanism regularizes prototype organization by capturing shared distributional structures across subjects. Furthermore, an adaptive prototype update strategy stabilizes prototype learning, and a relative consistency learning strategy reformulates classification as a similarity estimation problem to improve robustness against label noise.} Importantly, the target domain remains completely unseen during training, eliminating dependence on target data. The major contributions of this work are summarized as follows.}

\begin{itemize}
\color{black}
    \item We propose a unified framework (\textbf{MAT}) that disentangles domain and class representations, conceptualizing inter-subject EEG variability as a feature shift between domain-invariant and class-invariant components.
    \item We design a HDA mechanism with adaptive prototype updating to learn domain and class prototypes, effectively capturing shared cross-domain structures and enhancing generalization to unseen subjects.
    \item \textcolor{black}{We conduct extensive cross-subject validation on four public EEG datasets. Despite the absence of target-domain data, MAT outperforms or matches state-of-the-art transfer learning models and has been successfully extended to clinical mental illness detection scenarios.}
\end{itemize}
% Developed a graph contrastive learning method that generates positive samples through data augmentation and utilizes graph convolutional networks to capture complex relationships among multiple EEG channels, enhancing the effectiveness of graph-based feature representation.

\section{Related Work}
\label{sec:related_work}
\subsection{Non-Deep-Learning-Based Emotion Recognition}
\textcolor{black}{\indent Early studies on EEG-based emotion recognition primarily employed conventional signal processing and shallow machine learning methods to extract discriminative features and classify emotional states. These works provided foundational insights for the development of aBCIs. For example, Duan \etal \cite{Duan_Zhu_Lu_2013} proposed \textit{Differential Entropy} (DE) as an effective EEG feature, revealing that gamma-band activity correlates more strongly with emotional states than other frequency bands. Building on this work, Mohammadi \etal \cite{Mohammadi2017} employed well-established handcrafted features and traditional classifiers to identify emotional states from reduced frontal electrodes, achieving efficient recognition with limited channels. Liu \etal \cite{Yi-Hung_2013} presented the Kernel Eigen-Emotion Pattern (KEEP) method and an adaptive SVM to handle class imbalance, achieving higher accuracy than standard frequency-band power features. More recently, Luo \etal \cite{luo2024m3d} emphasized the joint consideration of marginal and conditional distributions during domain alignment, enabling dynamic adaptation between source and target domains to enhance cross-subject generalization. Despite these advances, non-deep-learning methods depend heavily on handcrafted feature engineering and static statistical assumptions, which constrain their robustness and limit generalization across subjects with varying physiological and emotional characteristics.}
%早期基于脑电信号的情感识别研究主要采用传统的信号处理和浅层机器学习方法提取具有判别性的特征并对情感状态进行分类。这些工作为aBCIs的开发提供了基础性的见解。例如，Duan \etal\cite{Duan_Zhu_Lu_2013}提出了\textit{差分熵}（DE）作为一种有效的脑电特征，揭示了伽马频带活动与情绪状态的相关性比其他频带更强。在这项工作的基础上，Mohammadi \etal\cite{Mohammadi2017}采用了完善的手工特征和传统分类器来从减少的额电极中识别情感状态，实现了有限通道的高效识别。Liu \etal\cite{Yi-Hung_2013}提出了核特征情感模式（Kernel Eigen-Emotion Pattern， KEEP）方法和自适应支持向量机（adaptive SVM）处理类不平衡问题，取得了比标准频带功率特征更高的准确率。最近，Luo \etal\cite{luo2024m3d}强调在域对齐过程中联合考虑边缘分布和条件分布，使源域和目标域之间的动态适应以增强跨主题泛化。尽管取得了这些进展，非深度学习方法在很大程度上依赖于手工特征工程和静态统计假设，这限制了它们的鲁棒性，并限制了对具有不同生理和情感特征的受试者的泛化。
%-------------------------------------------------------------------------
%--------------------------------------------------------------------------
%-------------------------------------------------------------------------
\begin{figure*}[th]
\centering
\subfloat{\includegraphics[width=0.9\textwidth]{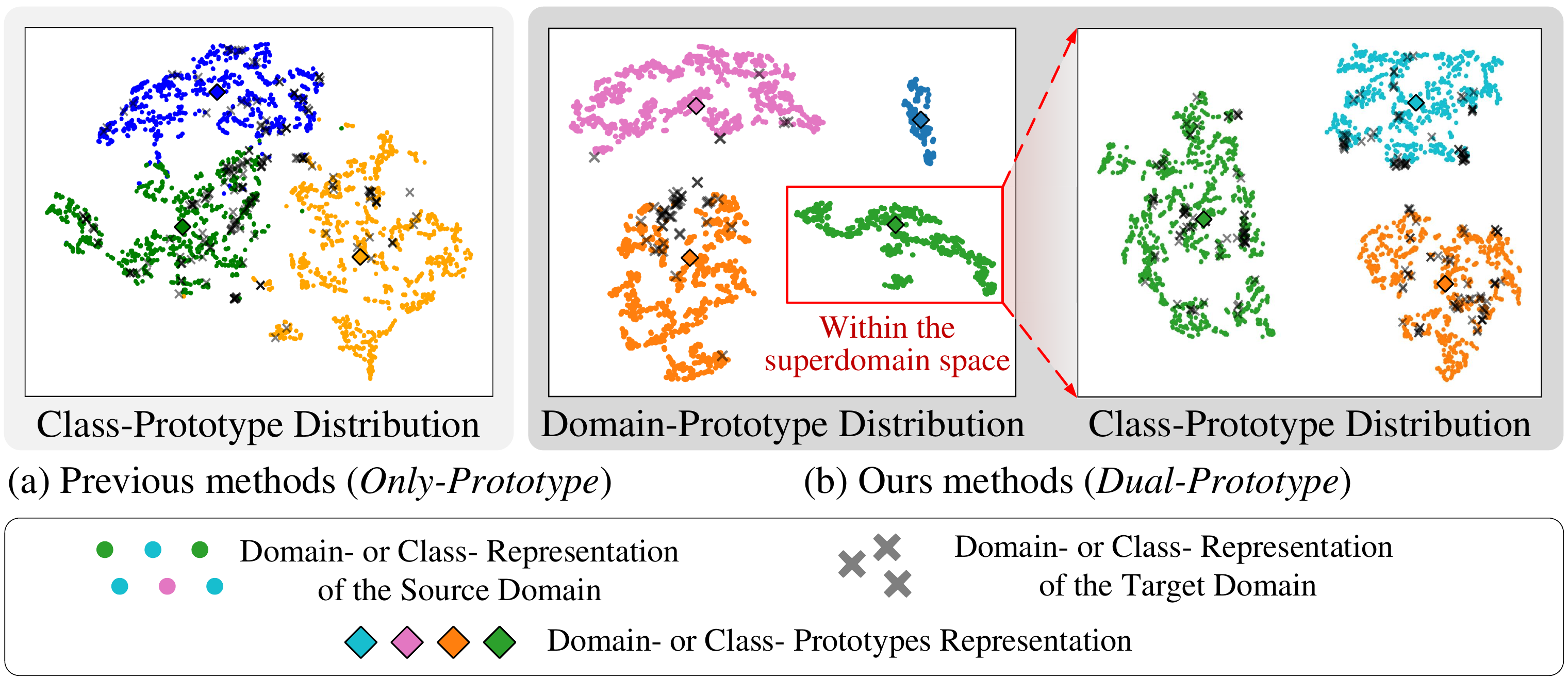}}
\caption{\textcolor{black}{The T-SNE visualization of the previous prototype representation method and ours. (a) Single class-prototype representation methods. (b) Our proposed dual-prototype representation method, which utilizes both domain-specific and class-specific prototype representations, effectively improves the generalization performance of the model.}}%迁移学习策略的基本框架，
\label{fig:Dual_P}
\end{figure*}
%--------------------------------------------------------------------------
\subsection{Deep Transfer Learning for Emotion Recognition}
\textcolor{black}{\indent Traditional EEG-based emotion recognition methods, though effective in small-scale or controlled settings, often struggle to generalize across subjects due to handcrafted feature limitations and rigid statistical assumptions. \textcolor{black}{To overcome these constraints, researchers have increasingly turned to deep learning techniques, which automatically learn hierarchical representations and enable knowledge transfer between subjects or experimental sessions. For example, Li \etal \cite{Li2025GNN} proposed a hierarchical dynamic local–global graph representation learning method (HDGNet) for EEG emotion recognition. Ding \etal \cite{ding2025emt} introduce a novel emotion transformer (EmT) model for generalized cross-subject emotion classification and regression. With the rapid development of deep learning, numerous transfer learning frameworks have been proposed to improve feature representation and cross-subject generalization. For example, Ye \etal \cite{Ye2023AdaptiveSA} proposed an emotion recognition model (GraphEmotionNet), which combined graph attention with transfer learning to enhance EEG feature representation. Zhou \etal \cite{zhou2023eeg} proposed an EEG-based Emotion Style Transfer Network (E2STN), integrating multi-domain EEG information to learn stylized emotional representations. Furthermore, to address the lable scarcity problem, Ye \etal \cite{ye2024semi} developed a graph-based contrastive learning framework (DS-AGC), while Liang \etal \cite{liang2025nssi} proposed a semi-supervised adversarial network (NSSI-Net), both of which effectively alleviate cross-domain distribution discrepancies under limited supervision.}
However, most existing deep transfer learning models require both source and target domain data during training, which limits their practicality in real-world scenarios where target data are unavailable or costly to collect.}
%Li \etal\cite{Li_Qiu_2020}提出了一种联合域适应网络，在对齐源域和目标域之间的潜在表示的同时，最小化源域的分类错误。Zhou \etal\cite{zhou2023eeg}提出了一种基于脑电的情感风格迁移网络（E2STN），融合源域的内容信息和目标域的风格信息形成风情化的情感表示。
%此外，半监督方法也取得了较大进步。叶\etal\cite{ye2024semi}开发了一种基于图的对比学习框架(DS-AGC)，Liang \etal\cite{liang2025nssi}提出了半监督对抗网络(NSSI-Net), 这有效缓解了标签缺失下的跨领域分布差异。
%然而，大多数现有的深度迁移学习模型在训练过程中同时需要源域和目标域数据，这限制了它们在目标数据不可用或收集成本高昂的现实场景中的实用性。

\subsection{Prototype Learning-Based Approaches}
\textcolor{black}{\indent Complementary to transfer learning, prototype learning has emerged as another promising paradigm for EEG-based emotion recognition. It represents each class with a representative feature vector, or \textit{prototype}, and performs classification by measuring the similarity between samples and these prototypes. This approach provides an interpretable and compact feature representation that can capture both global structure and intra-class consistency. Building on this concept, Zhou \etal \cite{zhou2023pr} proposed the Prototypical Representation-based Pairwise Learning (PR-PL) framework, which enhances generalization and robustness to noisy labels by reformulating classification as a similarity estimation task between sample pairs. Guo \etal \cite{GUO_2024} further unified prototype learning and adaptive learning theories in the Emotion Neural Network (EmNN), demonstrating that prototype-based mechanisms can effectively model high-level affective and behavioral representations across modalities. More recently, Zhou \etal \cite{zhou2024eegmatch} introduced EEGMatch, which jointly integrated prototype-wise and instance-wise representations to capture both global and local intrinsic relationships within EEG data.} \textcolor{black}{
These advances have significantly improved interpretability and robustness. However, existing methods generally learn only class prototypes and implicitly assume that all subjects follow a unified feature distribution, thereby overlooking domain-specific variations. As shown in Fig.~\ref{fig:Dual_P} (a), the single class-prototype representation leads to feature drift and blurred decision boundaries caused by individual differences, ultimately limiting cross-subject generalization. In contrast (Fig.~\ref{fig:Dual_P} (b)), the proposed dual-prototype mechanism effectively alleviates subject-specific feature drift and better reflects the underlying generative structure of EEG signals through complementary modeling of subject-specific and emotion-specific information.
%These advances have significantly improved interpretability and robustness. However, previous methods only class prototypes and implicitly assumes that all subjects follow a unified distribution, ignoring domain-specific variations. As shown in Fig.~\ref{fig:Dual_P} (a), the single class-prototype method makes the class features shift and fuzzy boundaries due to the individual (domain) differences, which ultimately limits the generalization performance. As shown in Fig.~\ref{fig:Dual_P} (b), our proposed dual-prototype mechanism effectively eliminates feature drift in specific subjects (domain) and better reflects the generation structure of EEG signals.
}
%这些进展显著提高了可解释性和鲁棒性.然而，以前的方法仅对类原型进行建模，并隐含地假设所有被试遵循统一的分布，这忽略了特定于领域的变化。如图1(a)所示，类特征因个体差异性进而导致特征偏移和边界模糊，最终限制了泛化性能。

%------------------------------------------------------------------------------
\begin{figure*}[ht]
\centering
\subfloat{\includegraphics[width=1\textwidth]{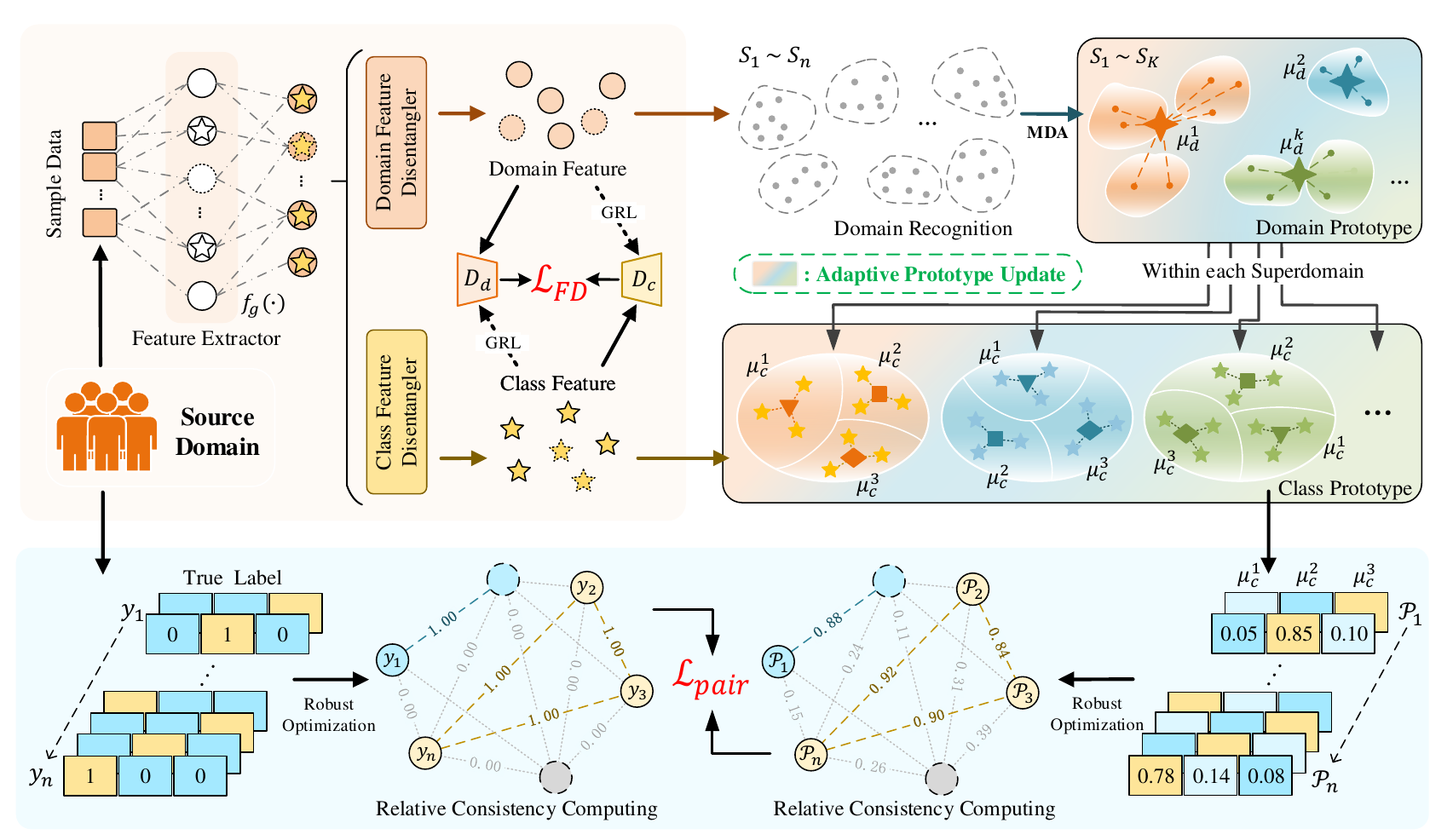}}
\caption{\textcolor{black}{The training phase of the MAT framework. Here, EEG features are first decoupled into domain- and class-specific components through discriminators $D_d$ and $D_c$ with a Gradient Reversal Layer (GRL) for adversarial alignment. Second, domains $S_1\sim S_n$ are clustered into superdomains $S_1\sim S_K$ by HDA mechanism to capture shared yet distinct subject representations. Within each superdomain, domain prototypes $\mu_d$ and class prototypes $\mu_c$ are adaptively updated to ensure stable and discriminative feature learning. Finally, the traditional classification task is transformed into the relative consistency learning between samples, improving robustness against label noise and unseen target domains.}}
\label{fig:MAT-Train}
\end{figure*}
%------------------------------------------------------------------------------
\begin{figure*}[ht]
\centering
\subfloat{\includegraphics[width=1\textwidth]{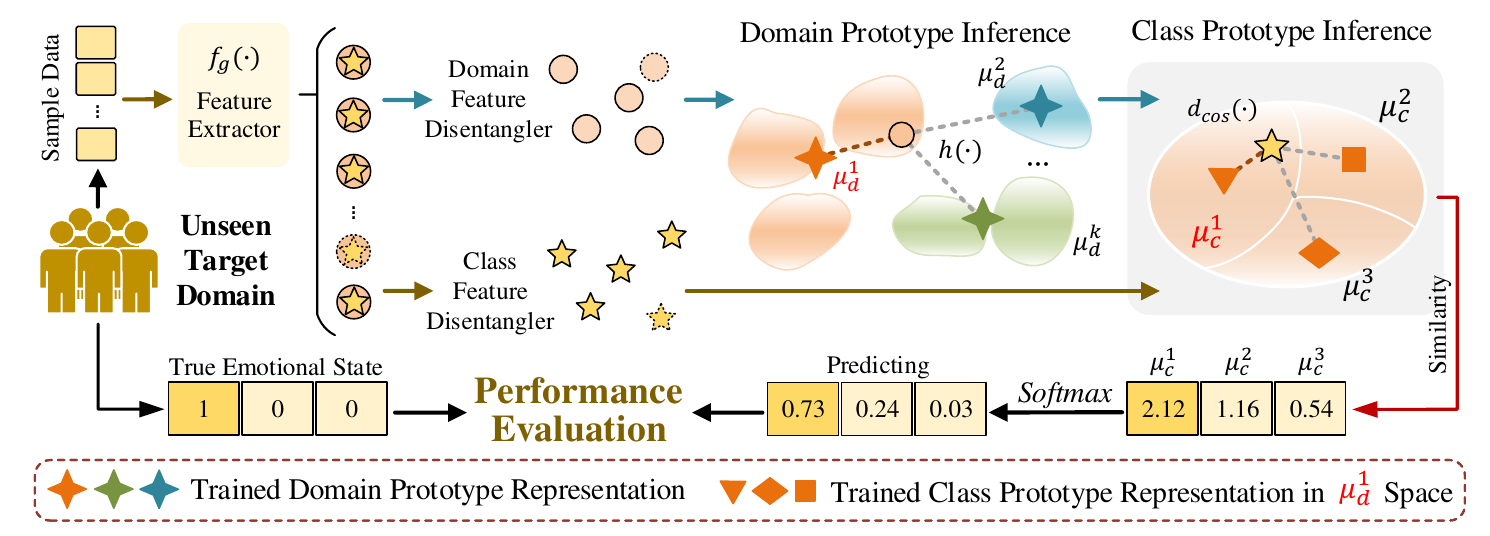}}
\caption{\textcolor{black}{The inferencing phase of the MAT framework. The optimal domain prototype $\mu_d$ and class prototype $\mu_c$ obtained during the model training phase are transferred to the unseen target domain. Specifically, we first decouple the sample features, and then the model determines the superdomain space $\mu_d$ by Domain Prototype Inference, and perform Class Prototype Inference within this superdomain space to determine the emotion category $\mu_c$. Here, $h(\cdot)$ represent the bilinear transformation to capture the most relevant domain space (Eq.\ref{Eq:Domain Inference}). $d_{cos}(\cdot)$ represents the similarity evaluation (Eq.\ref{Eq:Class Inference}).}}
\label{fig:MAT-inference}
\end{figure*} 
%------------------------------------------------------------------------------
%------------------------------------------------------------------------------
\begin{table}[]
\centering
\caption{Frequently used notations and descriptions.}
\label{tab:T1}
\scalebox{1}{
\begin{tabular}{cc}
\bottomrule
\textit{Notation}           & \textit{Description} \\
\midrule
$\mathbb{S} / \mathbb{T}$   & Source/Target Domain \\
$x_d / x_c $                & Domain-/Class-specific Representation  \\
$y_d / y_c$                 & Domain/Class Label  \\
$f_g(\cdot)$                & Shallow Feature Extractor  \\
$f_d(\cdot) / f_c(\cdot)$   & Domain/Class Feature Decoupler \\
$D_d(\cdot) / D_c(\cdot)$   & Domain / Class Discriminator \\ 
$\mu_d / \mu_c$             & \ Domain / Class Prototype Representation\ \\
$R(\cdot)$                  & Gradient Reversal Layer \\ 
$\mathcal{HK}$              & \textcolor{black}{Reproducing Kernel Hilbert Space} \\
$\kappa $                   & Gaussian Kernel Function \\
$\theta$                    & Bilinear Transformation Matrix \\
$K$                         & Number of Superdomains  \\
\bottomrule
\end{tabular}
}
\end{table}
%------------------------------------------------------------------------------

\section{Methodology}
\label{sec:Methodology}

\subsection{Feature Disentanglement and Representation Learning}
\label{Feature Disentanglement and Representation Learning}
\textcolor{black}{\indent Motivated by physiological and cognitive evidence, we assume that EEG signals comprise two complementary components: domain-invariant class features and class-invariant domain features. Individual differences are modeled as superimposed shifts of domain features on class features. Accordingly, EEG representations are formulated as the fusion of these two complementary subspaces, which are explicitly disentangled to improve cross-subject generalization.
%EEG signals contain two decoupled components: domain-invariant class features and class-invariant domain features. Individual differences are modeled as superimposed shifts of domain features with class features. We regard EEG features as a fusion of these two subspaces and aim to disentangle them for improved generalization across unseen subjects.
}
\textcolor{black}{Let the source and target domains be represented as $(\mathbb{S},\mathbb{T})$. In the source domain, each subject forms an independent subdomain:
\begin{equation}
    \mathbb{S} = \{S_n\}_{n=1}^{N_d}, \quad S_n = \{(x_n^i, y_n^i)\}_{i=1}^{N_s},
\end{equation}
where $x_n^i$ and $y_n^i$ denote the $i_{th}$ EEG sample and its corresponding emotion label from the $n_{th}$ subject. The target domain $\mathbb{T} = \{x_t^i, y_t^i\}_{i=1}^{N_t}$ remains entirely unseen during training. All notations are summarized in Table~\ref{tab:T1}.}
\textcolor{black}{As illustrated in Fig.~\ref{fig:MAT-Train}, a shallow feature extractor $f_g(\cdot)$ first captures latent EEG representations, which are then decomposed by a \textit{domain feature decoupler} $f_d(\cdot)$ and a \textit{class feature decoupler} $f_c(\cdot)$:
\begin{equation}
\label{xd_xc}
    x_d = f_d(f_g(x)), \quad x_c = f_c(f_g(x)),
\end{equation}
where $x_d$ and $x_c$ denote the domain-specific and class-specific representations, respectively.}
\textcolor{black}{To enforce proper separation of these two feature spaces, a domain discriminator $D_d(\cdot)$ and a class discriminator $D_c(\cdot)$ are adversarially trained through Gradient Reversal Layers (GRL). During backpropagation, GRL reverses gradients, thus encouraging each decoupler to learn representations invariant to the opposing factor. \textcolor{black}{
Before class-specific features are fed into the domain discriminator and domain-specific features into the class discriminator, the corresponding cross-task branches pass through a gradient reversal layer (GRL). The GRL acts as an identity mapping during forward propagation but multiplies the gradient propagated to the feature extractors by a negative coefficient during backpropagation. Therefore, $D_c$ and $D_d$ are trained to minimize their respective prediction losses, whereas $f_d$ and $f_c$ receive reversed gradients from the cross-task branches. This adversarial mechanism encourages $x_c$ to retain emotion-discriminative information while suppressing domain information and encourages $x_d$ to retain domain-related variations while suppressing emotion information.
} We employ the Binary Cross-Entropy loss ($\ell_{BCE}(\cdot)$) to optimize both discriminators, where the multi-class task is decomposed into independent binary classification objectives. Let $R(\cdot)$ represent the GRL operation. The class discriminator loss is defined as:
\begin{equation}
\label{Eq:L_cls}
    \mathcal{L}_{cls} = \ell_{BCE}[D_c(x_c), y_c] + \ell_{BCE}[D_c(R(x_d)), y_c],
\end{equation}
where $y_c$ is the true emotion label. Similarly, the domain discriminator loss is:
\begin{equation}
\label{Eq:L_dom}
    \mathcal{L}_{dom} = \ell_{BCE}[D_d(x_d), y_d] + \ell_{BCE}[D_d(R(x_c)), y_d],
\end{equation}
where $y_d$ denotes the domain (subject) label. 
\textcolor{black}{
The overall objective for the disentanglement module is given by:
\begin{equation}
\label{Eq:L_FD}
    \mathcal{L}_{FD} = \alpha\mathcal{L}_{cls} + \beta\mathcal{L}_{dom},
\end{equation}
where $\alpha=\beta=1$ in all experiments. These values were used because the class- and domain-disentanglement objectives are treated as equally important. All trainable components are jointly optimized in an end-to-end manner, without alternating updates. The detailed training procedures are presented in Supplementary Materials: Appendix D.}
}
\subsection{Hierarchical Domain Aggregation and Prototype Alignment}
\label{Hierarchical Domain Aggregation}

\textcolor{black}{\indent Each subject domain contains its unique data distribution. However, with a number of subjects, the model may overfit domain-specific biases and suffer efficiency loss. To address this problem, we design a HDA mechanism that merges related subject domains into higher-level \textit{superdomains}, followed by adaptive prototype alignment to capture the intrinsic cross-domain structure.}
\textcolor{black}{To quantify inter-domain relationships, we employ the Maximum Mean Discrepancy (MMD) as a statistical measure of distributional discrepancy. MMD is a non-parametric kernel-based metric that effectively captures nonlinear distribution differences in high-dimensional feature space. Given two domain feature sets $X \sim P$ and $Y \sim Q$, MMD is formulated as:
\begin{equation}
\label{Eq:MMD_HK}
    MMD^2_{\mathcal{HK}}(P,Q) = \left\| \mathbb{E}_x[\phi(X)] - \mathbb{E}_y[\phi(Y)] \right\|^2_{\mathcal{HK}},
\end{equation}
where $\phi(\cdot)$ denotes the feature mapping induced by the kernel function $\kappa(\cdot,\cdot)$ in the Reproducing Kernel Hilbert Space (RKHS). Considering the nonlinear and non-stationary \textcolor{black}{characteristics} of EEG signals, we employ a Gaussian kernel to implicitly project features into an infinite-dimensional RKHS:
\begin{equation}
\label{Eq:Kappa}
    \kappa(x,y) = \exp\left(-\frac{\|x-y\|^2}{2\sigma^2}\right),
\end{equation}
where $\sigma$ is the kernel bandwidth. The empirical estimation of MMD between two domains $X=\{x_i\}_{i=1}^{n}$ and $Y=\{y_j\}_{j=1}^{m}$ can then be expressed as:
\begin{equation}
\label{Eq:MMD^2_HK}
    \begin{split}
       MMD_{\mathcal{HK}}^2(X,Y)=
       \frac{1}{n(n-1)} \sum_{i=1}^{n} \sum_{j=1}^{n}\kappa(x_i,x_j)
       \\ + \frac{1}{m(m-1)} \sum_{i=1}^{m} \sum_{j=1}^{m}\kappa(y_i,y_j)
       -\frac{2}{nm} \sum_{i=1}^{n} \sum_{j=1}^{m}\kappa(x_i,y_j)   
    \end{split}
\end{equation}}

\textcolor{black}{By computing the pairwise MMD distances between all subject domains, we obtain an MMD distance vector for each domain:
\begin{equation}
\label{Eq:distance vector}
    \mathcal{V}_i = [MMD_{i,1}^2, MMD_{i,2}^2, \ldots, MMD_{i,N}^2],
\end{equation}
where $N$ denotes the number of domains. $\mathcal{V}_i$ encodes the relational distribution pattern of $i_{th}$ domain within the source domain space. Based on these distance vectors, we cluster related domains into superdomains using an MMD-based K-means++ algorithm, where the conventional Euclidean distance is replaced by MMD to better reflect inter-domain similarity:
\begin{equation}
\label{Eq:clustering algorithm}
    \mathcal{J} = \sum_{k=1}^{K}\sum_{(X,Y)\in S_k} MMD^2(X,Y),
\end{equation}
with $K$ represent the number of resulting superdomains. Consequently, the source domain $\mathbb{S}$ is reorganized into a hierarchical structure $\mathbb{S} = [S_1, S_2, \ldots, S_K]$, enabling the model to leverage both intra- and inter-domain relationships efficiently. \textcolor{black}{More HDA mechanism details are in Supplementary Materials: Appendix A.}}
\textcolor{black}{Within each superdomain $S_k$, we establish domain and class prototype representations to summarize the essential characteristics of the latent features. The domain prototype $\mu_d^k$ captures the centroid of domain-specific features within the $k$-th superdomain, while the class prototype $\mu_c^m$ captures the centroid of emotion-specific features for category $m$:
\begin{equation}
\label{Eq:mu_d^k}
    \mu_d^k = \frac{1}{|X_d^k|}\sum_{x_d^i\in X_d^k}x_d^i, \quad
    \mu_c^m = \frac{1}{|X_c^m|}\sum_{x_c^i\in X_c^m}x_c^i.
\end{equation}
These prototypes act as compact semantic anchors, representing both domain and class characteristics in a unified embedding space.}
\textcolor{black}{During model training, feature distributions continuously evolve, which can cause instability in prototype estimation. To ensure stable optimization and temporal coherence, we introduce an adaptive prototype alignment mechanism that updates prototype representations across iterations:
\begin{equation}
\label{Eq:mu_d^t}
    \mu_d^{(t)}=(1-\alpha)\mu_d^{(t-1)}+\alpha\mu_d^{(t)}, \quad
    \mu_c^{(t)}=(1-\alpha)\mu_c^{(t-1)}+\alpha\mu_c^{(t)},
\end{equation}
where $\alpha$ is a dynamic update coefficient that gradually decreases, which is  defined as:
\begin{equation}
\label{Eq:weight update parameter}
    \alpha = \alpha_l + (\alpha_h - \alpha_l)\!\left(1-\frac{t}{maxEpoch}\right)^p,
\end{equation}
with $\alpha_h$ and $\alpha_l$ denoting the upper and lower bounds and $p$ controlling the decay rate. This update mechanism enables rapid prototype adaptation in early training and smoother convergence in later stages as the feature space stabilizes.}

\textcolor{black}{Overall, the feature disentanglement, prototype learning, and hierarchical domain aggregation modules are jointly and interactively optimized.
%As shown in Fig.~2 in the Supplementary Materials, we visualize the interaction and joint optimization among feature disentanglement, prototype learning, and hierarchical domain aggregation. 
These three modules are connected through top-down prior constraints and bottom-up iterative optimization. Specifically, feature disentanglement defines the representation space for prototype learning, prototype optimization further refines the disentangled representations, and HDA regularizes prototype organization by capturing shared structures across subjects. Together, these three modules form a unified optimization framework for robust cross-subject representation learning. \textcolor{black}{More details are in Supplementary Materials: Appendix B.}}

\subsection{Inference and Noise-Robust Optimization}
\label{Inference and Noise-Robust Optimization}

\textcolor{black}{\indent During training, the proposed MAT framework learns domain and class prototypes within each superdomain through adaptive updating, while the target domain remains entirely unseen. In the inference phase (Fig.~\ref{fig:MAT-inference}), given a target EEG sample, the model first decouples its domain and class features using the trained feature extractors. The domain prototype representations learned in the source phase are then used to infer the most relevant superdomain for the target sample, followed by emotion classification through prototype-based matching within the identified superdomain.} 

\textcolor{black}{
To quantify the compatibility between a target-domain representation $x_d$ and the $k$-th domain prototype $\mu_d^k$, we define the bilinear score 
\begin{equation}
\label{Eq:bilinear transformation}
    h(x_d,\mu_d^k) = x_d^{\top}\theta\mu_d^k,
\end{equation}
where $\theta$ is a trainable bilinear transformation matrix. The matrix is randomly initialized and jointly optimized with the other trainable components by minimizing the overall objective $\mathcal{L}_{MAT}$. More specifically, $\mathcal{L}_{pair}$ provides direct supervision for $\theta$, with gradients propagated through the routing scores in Eq. (14). By contrast, $\mathcal{L}_{FD}$ does not directly update $\theta$ but improves the domain representations and prototypes used by the routing function. No symmetry or positive-definiteness constraint is imposed on $\theta$. During inference, the learned $\theta$ is fixed and used to route each unseen target sample to the most compatible superdomain, after which the corresponding class prototypes are used for emotion prediction.
%where ${\top}$ represents the transposition operation. $\theta$ is continuously updated during gradient backpropagation, and the optimization process of $\theta$ is evaluated by monitoring the loss trend of the final objective function.In addition, $\theta$ directly affects the hyperdomain aggregation in the training phase as well as the hyperdomain routing in the inference phase, and further affects the assignment of downstream class prototypes. If we replace $\theta$ with the identity matrix, it will directly affect the overall performance of the model, which is also confirmed by the ablation experiments.
}
The superdomain affiliation of each target sample is determined through a softmax normalization over all bilinear responses:
\begin{equation}
\label{Eq:Domain Inference}
    \mathcal{V}_i = softmax[h(x_d^i,\mu_d^1), \ldots, h(x_d^i,\mu_d^K)],
\end{equation}
where the index with the highest response corresponds to the predicted superdomain. Within this inferred superdomain, the emotion category is determined by evaluating the cosine similarity between the class feature $x_c^i$ and all class prototypes $\mu_c^m$:
\begin{equation}
\label{Eq:Class Inference}
    \mathcal{P}_i = softmax[d_{cos}(x_c^i,\mu_c^1), \ldots, d_{cos}(x_c^i,\mu_c^M)],
\end{equation}
where $d_{cos}(\cdot)$ denotes cosine similarity. The emotion label is assigned to the class with the highest similarity score. This hierarchical inference process enables the model to perform cross-subject generalization without accessing target labels.

\textcolor{black}{To further improve robustness against label noise, we reformulate the classification task as a pairwise similarity estimation problem. Instead of learning from single samples, the model learns relative consistency across sample pairs. Given two class feature embeddings $(x_c^i, x_c^j)$, we define the pairwise computing loss as:
\begin{equation}
    \label{Eq:pairwise learning}
    \begin{split}
       &\mathcal{L}_{pair}=\frac{1}{(N_b)^2}\cdot \\ &\sum_{i,j\in N_b}^{}\left [ r_{ij}\log_{}{(\mathcal{G}(x_c^i,x_c^j))}-(1-r_{ij})\log_{}{(1-\mathcal{G}(x_c^i,x_c^j))}  \right ], 
    \end{split}
\end{equation}
where $r_{ij}=1$ if the two samples belong to the same emotion category and $r_{ij}=0$ otherwise. The similarity function $\mathcal{G}(\cdot)$ measures the probability that two samples share the same emotion:
\begin{equation}
    \mathcal{G}(x_i,x_j)=\frac{\mathcal{P}_i\cdot\mathcal{P}_j}{\|\mathcal{P}_i\|_2\|\mathcal{P}_j\|_2}.
\end{equation}
This pairwise formulation encourages the model to learn relative relations between samples, mitigating the adverse effects of noisy or ambiguous labels by leveraging inter-sample consistency.}
\textcolor{black}{Finally, the overall objective integrates all modules:
\begin{equation}
    \label{Eq:MAT loss}
        \mathcal{L}_{MAT}=\mathcal{L}_{FD}+\alpha\mathcal{L}_{pair}+\beta\mathcal{R},
\end{equation}
where $\mathcal{R}$ denotes soft regularization, $\alpha$ and $\beta$ are weighting factor. The proposed MAT framework unifies feature disentanglement, hierarchical domain aggregation, prototype alignment, and noise-robust optimization into an end-to-end learning paradigm, enabling reliable emotion recognition across unseen subjects through adaptive and semantically consistent representation learning.}

\section{Experimental Results}
\label{sec:experiment}

\subsection{Databases and Data Processing}
\label{Databases and Data Processing}

\textcolor{black}{\indent We evaluated the proposed MAT framework using four public EEG emotion recognition databases: SEED \cite{7104132}, SEED-IV \cite{zheng2018emotionmeter}, SEED-V \cite{Liu_2022_eeg} and SEED-VII \cite{Jiang2025EEG}.} \textcolor{black}{The SEED dataset contains EEG recordings from 15 subjects, each participating in three experimental sessions conducted on different days. Each session includes 15 trials associated with three emotional categories: positive, neutral, and negative. Emotional states were elicited using video stimuli while EEG signals were simultaneously recorded. The SEED-IV dataset consists of 15 subjects, each completing three experimental sessions on separate dates. Each session comprises 24 trials covering four emotional categories: neutral, sad, fear, and happy. The SEED-V dataset includes 16 subjects, with each subject completing three sessions. Each session contains 15 trials corresponding to five emotional categories: happy, neutral, sad, disgust, and fear.}
\textcolor{black}{The SEED-VII dataset contained 20 subjects, each participating in four recording sessions conducted on different dates. Each session consists of 20 trials covering seven emotion categories: happy, sad, fear, disgust, surprise, anger, and neutral.}
\\ \indent
\textcolor{black}{The EEG data were preprocessed following a standard pipeline. First, the signals were downsampled to 200 Hz, and non-neural artifacts such as electromyography (EMG) and electrooculography (EOG) were manually removed. The cleaned signals were then bandpass filtered between 0.3 Hz and 50 Hz. Next, the signals were segmented using a 1 s sliding window. Differential entropy (DE) features, which characterize the logarithmic energy distribution of EEG signals, were extracted from five frequency bands: Delta (1 to 3 Hz), Theta (4 to 7 Hz), Alpha (8 to 12 Hz), Beta (14 to 30 Hz), and Gamma (31 to 50 Hz). This resulted in 310 features for each EEG segment, corresponding to 5 frequency bands across 62 channels. Finally, a Linear Dynamic System (LDS) was applied to smooth the extracted features by exploiting the temporal dependency of emotional responses, thereby suppressing noise-related and emotion-irrelevant components \cite{5627125}. This preprocessing procedure follows commonly adopted settings in prior studies to ensure fair comparisons with existing methods \cite{zhou2023pr,ye2024semi}.}

%--------------------------------------------------------------------------
\begin{table}[t]%SEED(1)
\begin{center}
\caption{Cross-subject single-session LOSO cross-validation results on SEED dataset, expressed as (Acc\% ± Std\%). Here, '*' indicates the results are obtained by our own implementation. '\#' denoted as the model of the unseen target domain and  \textcolor{orange}{$\uparrow$} represents the gap between MAT and them.}
\label{tab:SEED1}
\scalebox{1}{
\begin{tabular}{lc|lc}
\toprule
Methods                     & $P_{acc}(\%)$ & 
Methods                     & $P_{acc}(\%)$ \\
\midrule
\multicolumn{4}{c}{\textit{Traditional machine learning methods}
                    \textcolor{orange}{($\uparrow$ 13.22\%)}} \\ 
\midrule
KNN*\cite{KNN1982}          & 55.26 ± 12.43 &
KPCA*\cite{KPCA1999}        & 48.07 ± 09.97\\
SVM*\cite{SVM}              & 70.62 ± 09.02 & 
SA*\cite{SA2013}            & 59.73 ± 05.40 \\
TCA*\cite{TCA2010}          & 58.12 ± 09.52 & 
CORAL*\cite{CORAL2016}      & \underline{71.48} ± 11.57 \\
GFK*\cite{GFK2012}          & 56.71 ± 12.29 & 
RF*\cite{breiman2001random} & 62.78 ± 06.60 \\
\midrule
\multicolumn{4}{c}{\textit{Deep transfer learning methods} 
                \textcolor{orange}{($\uparrow$ 2.78\%)}}\\ 
\midrule
DAN*\cite{He2018DAN}        & 82.54 ± 09.25  & 
DANN*\cite{ganin2016DANN}   & 81.57 ± 07.21  \\
DDC*\cite{DDC2014}          & 75.42 ± 10.15  & 
DCORAL*\cite{Dcoral2016}    & 82.90 ± 06.97  \\
DGCNN\cite{song2018eeg}     & 79.95 ± 09.02  & 
MMD \cite{2013Equivalence}  & 80.88 ± 10.10  \\
BiDANN\cite{li2018bi}       & 83.28 ± 09.60  & 
GCPL \cite{li2024gene}      & 80.47 ± 06.05 \\
EFDMs\cite{zhang2019cross}  & 78.40 ± 06.76 & 
MS-MDA\cite{chen2021ms}     & 77.65 ± 11.32 \\
\textcolor{black}{BiHDM}\#\cite{9105104_2021}        & 81.55 ± 09.74 & 
\textcolor{black}{RGNN}\#\cite{9091308_2022}         & \underline{81.92} ± 09.35 \\
\textcolor{black}{EEG-DG}\#*\cite{10609514_2024}     & 81.14 ± 07.78 & 
\textcolor{black}{DMMR}\#*\cite{wang2024dmmr}        & 79.36 ± 08.24 \\
\midrule
\multicolumn{3}{l}{\textbf{MAT}} 
                            &\textbf{84.70 ± 04.63}\\
\bottomrule
\end{tabular}
}
\end{center}
\end{table}
%------------------------------------------------------------------------
\begin{table}[t]%SEED_IV(1)
\begin{center}
\caption{Cross-subject single-session LOSO cross-validation results on SEED-IV dataset, expressed as (Acc\% ± Std\%). Here, '*' indicates the results are obtained by our own implementation. '\#' denoted as the model of the unseen target domain and  \textcolor{orange}{$\uparrow$} represents the gap between MAT and them.}
\label{tab:SEED_IV1}
\scalebox{1}{
\begin{tabular}{lc|lc}
\toprule
Methods                     & $P_{acc}(\%)$ & 
Methods                     & $P_{acc}(\%)$ \\
\midrule
\multicolumn{4}{c}{\textit{Traditional machine learning methods}
                    \textcolor{orange}{($\uparrow$ 15.44\%)}} \\ 
\midrule
KNN*\cite{KNN1982}          & 41.77 ± 09.53 & 
KPCA*\cite{KPCA1999}        & 29.25 ± 09.73 \\
SVM*\cite{SVM}              & 50.50 ± 12.03 & 
SA*\cite{SA2013}            & 34.74 ± 05.29 \\
TCA*\cite{TCA2010}          & 44.11 ± 10.76 & 
CORAL*\cite{CORAL2016}      & 48.14 ± 10.38 \\
GFK*\cite{GFK2012}          & 43.10 ± 09.77 & 
RF*\cite{breiman2001random} & \underline{52.67} ± 13.85 \\
\midrule
\multicolumn{4}{c}{\textit{Deep transfer learning methods}
                    \textcolor{orange}{($\uparrow$ 3.84\%)}}\\ 
\midrule
DAN*\cite{He2018DAN}        & 59.27 ± 14.45  & 
DANN*\cite{ganin2016DANN}   & 57.16 ± 12.61  \\
DCORAL*\cite{Dcoral2016}    & 56.05 ± 15.60  & 
DDC*\cite{DDC2014}          & 58.02 ± 15.14  \\
MS-MDA\cite{chen2021ms}     & 61.43 ± 15.71  & 
MMD\cite{2013Equivalence}   & 59.34 ± 05.48  \\
DGCNN\cite{song2018eeg}     & 52.82 ± 09.23 &
BiDANN\cite{li2018bi}       & 65.59 ± 10.39 \\
\textcolor{black}{EEG-DG}\#*\cite{10609514_2024}    & \underline{64.27} ± 11.24 & 
\textcolor{black}{DMMR}\#*\cite{wang2024dmmr}       & 62.60 ± 12.90 \\
\midrule
\multicolumn{3}{l}{\textbf{MAT}} & \textbf{68.11 ± 11.02}\\
\bottomrule
\end{tabular}
}
\end{center}
\end{table}
%----------------------------------------------------------------------
%--------------------------------------------------------------

\begin{table}[t]%SEED_V(1)
\begin{center}
\color{black}
\caption{Cross-subject single-session LOSO cross-validation results on SEED-V dataset, expressed as (Acc\% ± Std\%). Here, '*' indicates the results are obtained by our own implementation. '\#' denoted as the model of the unseen target domain and  \textcolor{orange}{$\uparrow$} represents the gap between MAT and them.}
\label{tab:SEED_V1}
\scalebox{1}{
\begin{tabular}{lc|lc}
\toprule
Methods                     & $P_{acc}(\%)$ & 
Methods                     & $P_{acc}(\%)$ \\
\midrule
\multicolumn{4}{c}{\textit{Traditional machine learning methods}
                    \textcolor{orange}{($\uparrow$ 8.30\%)}} \\ 
\midrule
KNN*\cite{KNN1982}          & 35.73 ± 07.98 & 
KPCA*\cite{KPCA1999}        & 35.47 ± 09.39 \\
SVM*\cite{SVM}              & 53.14 ± 10.10 & 
SA*\cite{SA2013}            & 36.06 ± 11.55 \\
TCA*\cite{TCA2010}          & 37.57 ± 13.47 & 
CORAL*\cite{CORAL2016}      & \underline{55.18} ± 07.42 \\
GFK*\cite{GFK2012}          & 38.32 ± 10.11 & 
RF*\cite{breiman2001random} & 42.29 ± 16.02 \\
\midrule
\multicolumn{4}{c}{\textit{Deep transfer learning methods}
                    \textcolor{orange}{($\uparrow$ 4.45\%)}}\\ 
\midrule
DAN*\cite{He2018DAN}        & 59.36 ± 16.83  & 
DANN*\cite{ganin2016DANN}   & 56.28 ± 16.25  \\
DCORAL*\cite{Dcoral2016}    & 56.26 ± 14.56  & 
DDC*\cite{DDC2014}          & 56.54 ± 18.35  \\
EEG-DG\#*\cite{10609514_2024}     & \underline{59.03} ± 12.14 & 
DMMR\#*\cite{wang2024dmmr}        & 58.63 ± 15.30 \\
\midrule
\multicolumn{3}{l}{\textbf{MAT}} & \textbf{63.48 ± 10.86}\\
\bottomrule
\end{tabular}
}
\end{center}
\end{table}
%--------------------------------------------------------------

\begin{table}[t]%SEED_VII(1)
\color{black}
\begin{center}
\caption{Cross-subject single-session LOSO cross-validation results on SEED-VII dataset, expressed as (Acc\% ± Std\%). Here, '*' indicates the results are obtained by our own implementation. '\#' denoted as the model of the unseen target domain and  \textcolor{orange}{$\uparrow$} represents the gap between MAT and them.}
\label{tab:SEED_VII1}
\scalebox{1}{
\begin{tabular}{lc|lc}
\toprule
Methods                     & $P_{acc}(\%)$ & 
Methods                     & $P_{acc}(\%)$ \\
\midrule
\multicolumn{4}{c}{\textit{Traditional machine learning methods}
                    \textcolor{orange}{($\uparrow$ 9.56\%)}} \\ 
\midrule
KNN*\cite{KNN1982}          & 21.72 ± 06.87 & 
KPCA*\cite{KPCA1999}        & 35.93 ± 09.81 \\
SVM*\cite{SVM}              & 31.21 ± 09.37 & 
SA*\cite{SA2013}            & 30.45 ± 11.31 \\
TCA*\cite{TCA2010}          & \underline{37.34} ± 08.82 & 
CORAL*\cite{CORAL2016}      & 27.96 ± 07.22 \\
GFK*\cite{GFK2012}          & 31.34 ± 09.66 & 
RF*\cite{breiman2001random} & 34.82 ± 08.17 \\
\midrule
\multicolumn{4}{c}{\textit{Deep transfer learning methods}
                    \textcolor{orange}{($\uparrow$ 2.21\%)}}\\ 
\midrule
DAN*\cite{He2018DAN}        & 42.34 ± 06.04  & 
DANN*\cite{ganin2016DANN}   & 38.56 ± 13.20  \\
DCORAL*\cite{Dcoral2016}    & 34.15 ± 09.32  & 
PL-DCP\#*\cite{LI2026DG}         & \underline{44.69} ± 06.71  \\
EEG-DG\#*\cite{10609514_2024}     & 43.61 ± 08.12 & 
DMMR\#*\cite{wang2024dmmr}        & 43.60 ± 11.30 \\
\midrule
\multicolumn{3}{l}{\textbf{MAT}} & \textbf{46.90 ± 07.72}\\
\bottomrule
\end{tabular}
}
\end{center}
\end{table}

%-----------------------------------------------------------
\begin{table}[t]%SEED(2)
\begin{center}
\caption{Cross-subject cross-session LOSO cross-validation results on SEED dataset, expressed as (Acc\% ± Std\%). Here, '*' indicates the results are obtained by our own implementation. '\#' denoted as the model of the unseen target domain and  \textcolor{orange}{$\uparrow$} represents the gap between MAT and them.} 
\label{tab:SEED2}
\scalebox{1}{
\begin{tabular}{lc|lc}
\toprule
Methods                     & $P_{acc}(\%)$ & 
Methods                     & $P_{acc}(\%)$ \\
\midrule
\multicolumn{4}{c}{\textit{Traditional machine learning methods}
                    \textcolor{orange}{($\uparrow$ 7.43\%)}} \\ 
\midrule
KNN*\cite{KNN1982}          & 60.18 ± 08.10 & 
KPCA*\cite{KPCA1999}        & 72.56 ± 06.41 \\
SVM*\cite{SVM}              & 68.01 ± 07.88 & 
SA*\cite{SA2013}            & 57.47 ± 10.01 \\
TCA*\cite{TCA2010}          & 63.63 ± 06.40 & 
CORAL*\cite{CORAL2016}      & 55.18 ± 07.42 \\
GFK*\cite{GFK2012}          & 60.75 ± 08.32 & 
RF*\cite{breiman2001random} & \underline{72.78} ± 06.60 \\
\midrule
\multicolumn{4}{c}{\textit{Deep transfer learning methods}
                    \textcolor{orange}{($\uparrow$ 2.48\%)}}\\ 
\midrule
DAN*\cite{He2018DAN}        & 78.12 ± 05.47 & 
DANN*\cite{ganin2016DANN}   & 78.42 ± 07.57 \\
DCORAL*\cite{Dcoral2016}    & 77.36 ± 06.27 &
DDC*\cite{DDC2014}          & 73.22 ± 05.48 \\
MS-MDA                      & 78.14 ± 06.18 &
EmT-B\cite{ding2025emt}     & 78.80 ± 12.00 \\
IAG\cite{song2020instance}  & 74.36 ± 11.11 &
AMDET\cite{xu2023amdet}     & 72.10 ± 16.80 \\
EmT-S\cite{ding2025emt}     & 78.00 ± 11.70 &
GCB-Net\cite{zhang2019gcb}  & 68.40 ± 17.20 \\
\textcolor{black}{Mixup}\#\cite{zhang2018mixup}      & \underline{77.73} ± 10.02 & 
\textcolor{black}{RSC-MLP2}\#\cite{huang2020self}    & 76.74 ± 08.52 \\
\textcolor{black}{EEG-DG}\#*\cite{10609514_2024}     & 73.29 ± 07.37 & 
\textcolor{black}{DMMR}\#*\cite{wang2024dmmr}        & 77.67 ± 08.91 \\
\midrule
\multicolumn{3}{l}{\textbf{MAT}} & \textbf{80.21 ± 06.32}\\
\bottomrule
\end{tabular}
}
\end{center}
\end{table}
%-----------------------------------------------------------
\begin{table}[t]%SEED_IV(2)
\begin{center}
\caption{Cross-subject cross-session LOSO cross-validation results on SEED-IV dataset, expressed as (Acc\% ± Std\%). Here, '*' indicates the results are obtained by our own implementation. '\#' denoted as the model of the unseen target domain and  \textcolor{orange}{$\uparrow$} represents the gap between MAT and them.}
\label{tab:SEED_IV2}
\scalebox{1}{
\begin{tabular}{lc|lc}
\toprule
Methods                     & $P_{acc}(\%)$ & 
Methods                     & $P_{acc}(\%)$ \\
\midrule
\multicolumn{4}{c}{\textit{Traditional machine learning methods}
                    \textcolor{orange}{($\uparrow$ 14.5\%)}} \\ 
\midrule
KNN*\cite{KNN1982}          & 40.06 ± 04.98 & 
KPCA*\cite{KPCA1999}        & 47.79 ± 07.85 \\
SVM*\cite{SVM}              & 48.36 ± 07.51 & 
SA*\cite{SA2013}            & 40.34 ± 05.85 \\
TCA*\cite{TCA2010}          & 43.01 ± 07.13 & 
CORAL*\cite{CORAL2016}      & \underline{50.01} ± 07.93 \\
GFK*\cite{GFK2012}          & 43.48 ± 06.27 & 
RF*\cite{breiman2001random} & 48.16 ± 09.43 \\
\midrule
\multicolumn{4}{c}{\textit{Deep transfer learning methods}
                    \textcolor{orange}{($\uparrow$ 3.23\%)}}\\ 
\midrule
DAN*\cite{He2018DAN}        & 60.95 ± 09.34 & 
DANN*\cite{ganin2016DANN}   & 61.44 ± 11.66 \\
DCORAL*\cite{Dcoral2016}    & 59.96 ± 09.03 & 
DDC*\cite{DDC2014}          & 54.76 ± 09.02 \\
GCPL\cite{li2024gene}       & 62.65 ± 09.79 &
MS-MDA\cite{chen2021ms}     & 59.34 ± 05.48 \\
A-LSTM\cite{song2019mped}   & 55.03 ± 09.28 &
IAG\cite{song2020instance}  & 62.64 ± 10.25 \\
ADAST\cite{eldele2022adast} & 53.66 ± 13.63 &
MFA-LR\cite{jimenez2023}    & 61.66 ± 11.53 \\
\textcolor{black}{EEG-DG}\#*\cite{10609514_2024}    & \underline{61.28} ± 11.25 & 
\textcolor{black}{DMMR}\#*\cite{wang2024dmmr}       & 58.26 ± 11.92 \\
\midrule
\multicolumn{3}{l}{\textbf{MAT}} & \textbf{64.51 ± 09.22} \\
\bottomrule
\end{tabular}
}
\end{center}
\end{table}
%-----------------------------------------------------------
\begin{table}[t]%SEED_V(2)
\begin{center}
\caption{Cross-subject cross-session LOSO cross-validation results on SEED-V dataset, expressed as (Acc\% ± Std\%). Here, '*' indicates the results are obtained by our own implementation. '\#' denoted as the model of the unseen target domain and  \textcolor{orange}{$\uparrow$} represents the gap between MAT and them.}
\label{tab:SEED_V2}
\scalebox{1}{
\begin{tabular}{lc|lc}
\toprule
Methods                     & $P_{acc}(\%)$ & 
Methods                     & $P_{acc}(\%)$ \\
\midrule
\multicolumn{4}{c}{\textit{Traditional machine learning methods}
                    \textcolor{orange}{($\uparrow$ 4.31\%)}} \\ 
\midrule
KNN*\cite{KNN1982}          & 35.28 ± 07.57 & 
KPCA*\cite{KPCA1999}        & 39.68 ± 11.28 \\
SVM*\cite{SVM}              & 41.20 ± 10.76 & 
SA*\cite{SA2013}            & 31.87 ± 09.87 \\
TCA*\cite{TCA2010}          & 37.68 ± 08.40 & 
CORAL*\cite{CORAL2016}      & \underline{54.08} ± 07.44 \\
GFK*\cite{GFK2012}          & 37.89 ± 09.84 & 
RF*\cite{breiman2001random} & 43.63 ± 11.38 \\
\midrule
\multicolumn{4}{c}{\textit{Deep transfer learning methods}
                    \textcolor{orange}{($\uparrow$ 3.23\%)}}\\ 
\midrule
DAN*\cite{He2018DAN}        & 54.27 ± 10.42 & 
DANN*\cite{ganin2016DANN}   & 52.83 ± 13.90 \\
DCORAL*\cite{Dcoral2016}    & 52.23 ± 12.76 & 
DDC*\cite{DDC2014}          & 43.89 ± 11.84 \\
ADAST\cite{eldele2022adast} & 50.69 ± 14.01 &
A-LSTM\cite{song2019mped}   & 40.34 ± 08.68 \\
\textcolor{black}{EEG-DG}\#*\cite{10609514_2024}   & 53.92 ± 13.25 & 
\textcolor{black}{DMMR}\#*\cite{wang2024dmmr}      & \underline{55.16} ± 12.71 \\
\midrule
\multicolumn{3}{l}{\textbf{MAT}} & \textbf{58.39 ± 08.63}\\
\bottomrule
\end{tabular}
}
\end{center}
\end{table}

%--------------------------------------------------------------

\begin{table}[t]%SEED_VII(2)
\color{black}
\begin{center}
\caption{Cross-subject cross-session LOSO cross-validation results on SEED-VII dataset, expressed as (Acc\% ± Std\%). Here, '*' indicates the results are obtained by our own implementation. '\#' denoted as the model of the unseen target domain and  \textcolor{orange}{$\uparrow$} represents the gap between MAT and them.}
\label{tab:SEED_VII2}
\scalebox{1}{
\begin{tabular}{lc|lc}
\toprule
Methods                     & $P_{acc}(\%)$ & 
Methods                     & $P_{acc}(\%)$ \\
\midrule
\multicolumn{4}{c}{\textit{Traditional machine learning methods}
                    \textcolor{orange}{($\uparrow$ 5.50\%)}} \\ 
\midrule
KNN*\cite{KNN1982}          & 19.03 ± 03.09 & 
KPCA*\cite{KPCA1999}        & \underline{28.99} ± 05.84 \\
SVM*\cite{SVM}              & 22.50 ± 04.70 & 
SA*\cite{SA2013}            & 17.98 ± 03.89 \\
TCA*\cite{TCA2010}          & 28.58 ± 07.74 & 
CORAL*\cite{CORAL2016}      & 20.44 ± 04.64 \\
GFK*\cite{GFK2012}          & 27.12 ± 06.25 & 
RF*\cite{breiman2001random} & 27.13 ± 04.44 \\
\midrule
\multicolumn{4}{c}{\textit{Deep transfer learning methods}
                    \textcolor{orange}{($\uparrow$ 2.82\%)}}\\ 
\midrule
DAN*\cite{He2018DAN}             & 29.20 ± 03.77 & 
DANN*\cite{ganin2016DANN}        & 25.51 ± 05.43 \\
DCORAL*\cite{Dcoral2016}         & 24.98 ± 04.41 & 
PL-DCP\#*\cite{LI2026DG}         & 30.17 ± 03.60 \\
EEG-DG\#*\cite{10609514_2024}    & 30.72 ± 05.12 & 
DMMR\#*\cite{wang2024dmmr}       & \underline{31.67} ± 05.28 \\
\midrule
\multicolumn{3}{l}{\textbf{MAT}} & \textbf{34.49 ± 04.22}\\
\bottomrule
\end{tabular}
}
\end{center}
\end{table}

%-----------------------------------------------------------
%----------------------------------------------------------------------------------

\subsection{Model Implementation}

\textcolor{black}{\indent In the proposed model, the architecture of the shallow feature extractor $f_g(\cdot)$ consists of an input layer with 310 units, followed by two fully connected hidden layers with 64 units each. LeakyReLU activations with a negative slope of $\alpha=0.01$ are applied after each hidden layer, and the output dimension is set to 64. The domain and class feature decouplers, denoted as $f_d(\cdot)$ and $f_c(\cdot)$, share the same network structure. Each decoupler includes an input layer of 64 units, two hidden layers of 64 units with ReLU activations, and an output layer of 64 units. Similarly, the domain and class discriminators $D_d(\cdot)$ and $D_c(\cdot)$ adopt identical architectures. Each discriminator consists of an input layer with 64 units, followed by a hidden layer with 64 units, a dropout layer with a dropout rate $p=0.25$, another hidden layer with 64 units, and a sigmoid-activated output layer with 64 units. In the Multi-Domain Aggregation module, the number of aggregated superdomains is set to 4. For the adaptive prototype updating strategy in Eq.~\ref{Eq:weight update parameter}, the upper and lower bounds of the update weight are set to $\alpha_h=0.8$ and $\alpha_l=0.2$, respectively, and the decay hyperparameter $p$ is set to 2. The weighting factor $\alpha$ and $\beta$ in Eq.~\ref{Eq:MAT loss} is fixed to 1.0 and 0.02. All experiments are conducted using Python 3.8 with CUDA 11.8 and PyTorch 2.0.0. Additional dependencies include NumPy 1.24.3 and Scikit-learn 0.22.1. All models are trained and evaluated on an NVIDIA GeForce RTX 3090 GPU.}

\subsection{Experiment Protocols}

\textcolor{black}{\indent To comprehensively evaluate the proposed model and facilitate fair comparisons with existing methods, we adopt two cross-validation protocols. In both settings, the target-domain data are completely unseen during training. \textbf{(1) Cross-Subject Single-Session Leave-One-Subject-Out (LOSO) Cross-Validation.} This protocol is widely used in EEG-based emotion recognition. In each fold, data from one subject are selected as the target domain, while data from the remaining subjects serve as the source domain. Following common practice in prior studies, only data from the first session of each subject are used. The training and evaluation process is repeated until each subject has been treated as the target domain, and the final performance is obtained by averaging results across all folds. \textbf{(2) Cross-Subject Cross-Session LOSO Cross-Validation.} To better reflect real-world application scenarios, all sessions of one subject are used as the target domain, and all sessions of the remaining subjects are used as the source domain. The procedure is repeated such that each subject is evaluated as the target domain once, and the results are averaged across subjects. This protocol introduces additional inter-session variability and represents a more challenging evaluation setting for EEG-based emotion recognition.}

\subsection{Cross-Subject Single-Session LOSO Cross-Validation}
\label{Single-Session}

\textcolor{black}{\indent We evaluate the proposed MAT model and compare it with representative state-of-the-art methods. The results on the SEED dataset are reported in Table~\ref{tab:SEED1}. MAT demonstrates clear performance advantages over traditional machine learning approaches. Specifically, MAT achieved a performance of 84.70\%, which was 13.22\% higher than the optimal machine learning method (CORAL, 71.48\%). Compared with the deep learning baseline, MAT achieved the best performance. The accuracy of the suboptimal model BiDANN is 83.28\%, and the MAT has increased by 1.42\%. In the baseline with unseen target domain, MAT has increased by 2.78\% compared with the suboptimal model (RGNN, 81.92\%). The evaluation results on the SEED-IV dataset are summarized in Table~\ref{tab:SEED_IV1}. In both machine learning and deep transfer learning methods, MAT has achieved the highest accuracy. In the baseline, RF and EEG-DG achieved suboptimal performance in machine learning and deep learning method with unseen target domain, with accuracy of 52.67\% and 64.27\% respectively. The performance of MAT was 68.11\%, which was improved by 15.44\% and 3.84\% respectively compared with these methods. \textcolor{black}{The results on the SEED-V dataset are presented in Table~\ref{tab:SEED_V1}. MAT achieves an accuracy of 63.48\%, exceeding the performance of all compared baseline methods. Compared with the suboptimal deep learning model with unseen target (EEG-DG, 59.03\%), MAT improves the accuracy by 4.45\%.}} 
\textcolor{black}{
The experimental results on the SEED-VII dataset are reported in Table~\ref{tab:SEED_VII1}. MAT achieves an accuracy of 46.90\%, outperforming the best unseen-target deep transfer learning baseline (PL-DCP, 44.69\%) by 2.21\% and the best traditional machine-learning baseline (TCA, 37.34\%) by 9.56\%. Unlike most competing methods, MAT does not require any target-domain data during training. These results demonstrate its superior domain generalization capability under unseen target scenarios.}

\subsection{Cross-Subject Cross-Session LOSO Cross-Validation}
\label{Cross-Session}

\textcolor{black}{\indent Compared with the cross-subject single-session setting, the cross-subject cross-session protocol further accounts for inter-session variability in addition to inter-subject differences. The evaluation results on the SEED dataset are reported in Table~\ref{tab:SEED2}. MAT achieves an accuracy of 80.21\%, outperforming both machine learning and deep learning baseline methods. Among the machine learning methods, RF achieves the best performance with an accuracy of 72.78\%, and MAT has an improvement of 7.43\%. Compared with the suboptimal deep learning methods with unseen target (Mixup, 77.73\%), MAT improves by 2.48\%. Notably, MAT achieves competitive performance without access to target-domain data during training. The results on the SEED-IV dataset are summarized in Table~\ref{tab:SEED_IV2}. MAT achieves the highest performance of 64.51\% among all compared baselines. Compared with the suboptimal baseline model (EEG-DG, 61.28\%), MAT improved performance by 3.23\%. The evaluation results on the SEED-V dataset are presented in Table~\ref{tab:SEED_V2}. MAT achieved a performance of 58.39\% in five-class emotion recognition. The suboptimal baselines CORAL and DMMR achieved accuracy of 54.08\% and 55.16\%, respectively. Compared with DMMR, MAT increased by 3.23\%.} 
\textcolor{black}{
The results on the more challenging cross-subject cross-session SEED-VII setting are reported in Table~\ref{tab:SEED_VII2}. MAT achieves an accuracy of 34.49\%, outperforming the strongest unseen-target baseline (DMMR, 31.67\%) by 2.82\%. These results further demonstrate that MAT effectively addresses both inter-subject and inter-session variability without requiring target-domain data during training.}

\section{Discussion}
\label{Discussion}

%--------------------------------------------------------------------
\begin{table}[t]
\centering
\caption{Results of ablation experiments of the MAT model, expressed as (Acc\% ± Std\%)}
\label{tab:Ablation}
\scalebox{1.12}{
\begin{tabular}{lcl}
\toprule
Ablation Strategy & $P_{acc}(\%)$ \\
\midrule
\textit{w/o Domain Prototype}
                                    & 78.95 ± 08.92 
                                    & \textcolor{orange}{$\downarrow$ 5.75\%}\\
\textit{w/o Class Disc. Loss in Eq.~\ref{Eq:L_cls}}   
                                    & 81.62 ± 07.16
                                    & \textcolor{orange}{$\downarrow$ 3.08\%}\\
\textit{w/o Domain Disc. Loss in Eq.~\ref{Eq:L_dom}}  
                                    & 79.34 ± 08.74
                                    & \textcolor{orange}{$\downarrow$ 5.36\%}\\
\textit{w/o Class and Domain Disc. Loss}              
                                    & 78.11 ± 09.24
                                    & \textcolor{orange}{$\downarrow$ 6.59\%}\\
\textit{w/o HDA mechanism in Sec.~\ref{Hierarchical Domain Aggregation}}                
                                    & 80.23 ± 05.12
                                    & \textcolor{orange}{$\downarrow$ 4.47\%}\\
\textit{w/o Ada. Upd. Coe. $\alpha$ in Eq.~\ref{Eq:mu_d^t}}                      
                                    & 81.54 ± 05.86
                                    & \textcolor{orange}{$\downarrow$ 3.16\%}\\
\textit{w/o Bil. Trans. Matrix $\theta$ in Eq.~\ref{Eq:bilinear transformation}}                                                              & 82.94 ± 06.31
                                    & \textcolor{orange}{$\downarrow$ 1.76\%}\\
\textit{w/o Pair. Computing in Eq.~\ref{Eq:pairwise learning}}                      
                                    & 76.73 ± 06.62 
                                    & \textcolor{orange}{$\downarrow$ 7.97\%}\\
\textit{w/o Soft Reg. $\mathcal{R}$ in Eq.~\ref{Eq:MAT loss}}                                                                          & 83.65 ± 04.89
                                    &\textcolor{orange}{$\downarrow$ 1.05\%}\\
\midrule
\textbf{\textbf{MAT}}                    
&\textbf{84.70 ± 04.63} \\
\bottomrule
\end{tabular}
}
\end{table}
%--------------------------------------------------------------------
% %------------------------------------------------------------------------------
% \begin{figure*}[t]
% \centering
% \includegraphics[width=1\textwidth]{./Figures/SEED_SEEDIV_1225}
% \caption{\textcolor{black}{Confusion matrices of different baseline models under cross-subject single-session LOSO cross-validation. The SEED database (a)$\sim$(d) contains three emotion categories: negative, neutral and positive. The Seed-IV database (e)$\sim$(h) contains four emotion categories: neutral, sad, fear and happy. The horizontal axis represents the predicted labels, while the vertical axis represents the true labels.}}
% \label{fig:SEED_SEEDIV}
% \end{figure*}
% %------------------------------------------------------------------------------
% \begin{figure*}[t]
% \centering
% \includegraphics[width=1\textwidth]{./Figures/SEEDV_1225}
% \caption{\textcolor{black}{Confusion matrices of different baseline models under cross-subject single-session LOSO cross-validation. The Seed-V database contains five emotion categories: happy, fear, neutral, sad, and disgust. The horizontal axis represents the predicted labels, while the vertical axis represents the true labels.}}
% \label{fig:SEEDV}
% \end{figure*}
% %-----------------------------------------------------------------------------

\subsection{Ablation Experiment}
\textcolor{black}{\indent To assess the contribution of each component in the proposed MAT framework, we conduct ablation experiments on the SEED dataset under the cross-subject single-session LOSO cross-validation protocol.}

\begin{figure*}[t]
\centering
\includegraphics[width=1\textwidth]{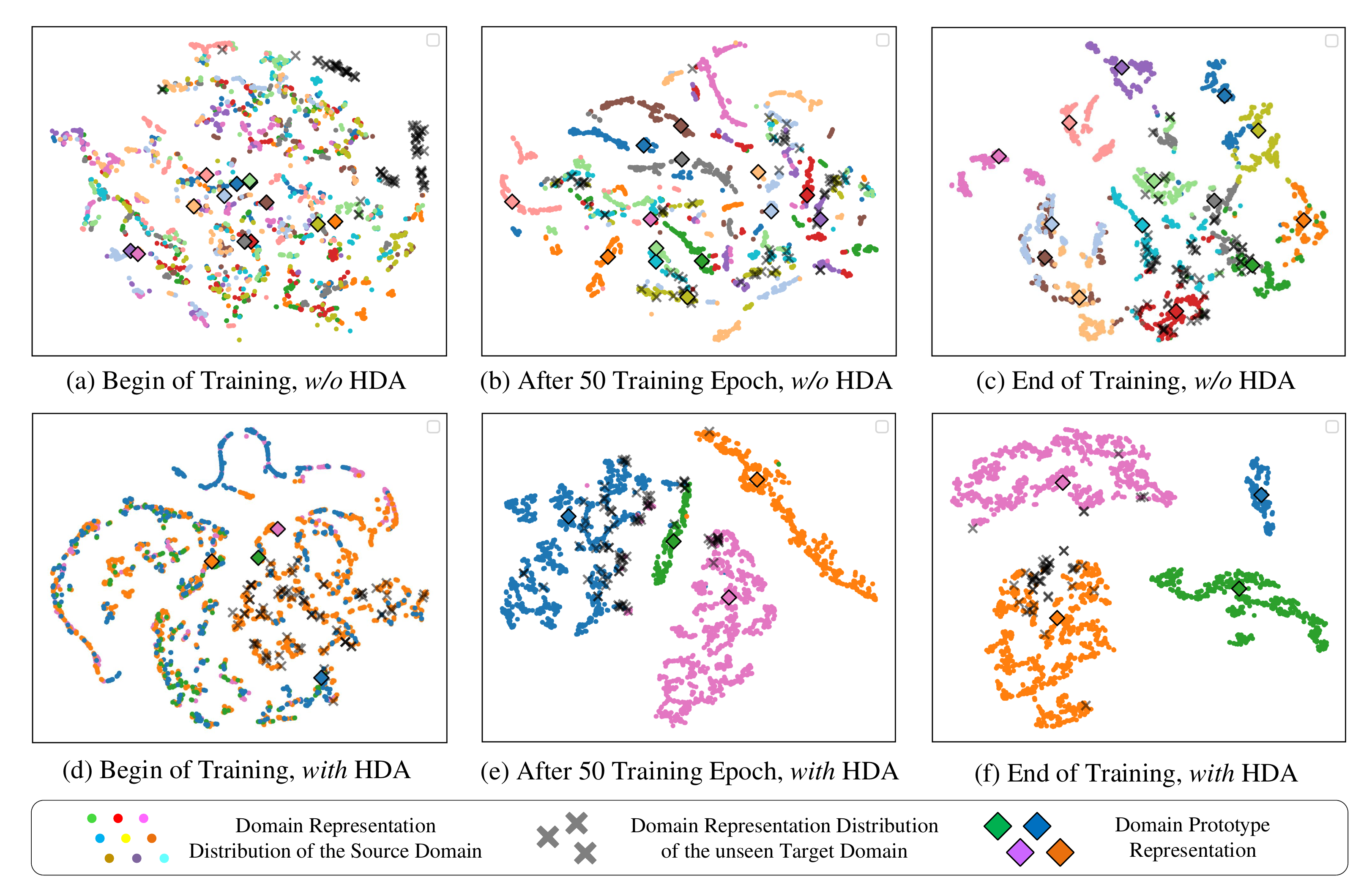}
\caption{\textcolor{black}{T-SNE visualization of domain representation and domain alignment \textit{without} (a) to (c) and \textit{with} (d) to (f) Hierarchical-Domain Aggregation mechanism. And showing the distribution of domain representations at the beginning of training, after 50 training epochs, and end of training, respectively.}}
\label{fig:Domain_feature}
\end{figure*}

%\textcolor{red}{The ablation results are reported in Table~\ref{tab:Ablation}. When the domain prototype representation is removed and only class prototypes are retained, the performance decreases by 5.75\%, indicating the importance of domain prototypes in capturing subject-related characteristics. Removing the class discriminator loss results in a performance drop from 84.70\% to 81.62\%, corresponding to a decrease of 3.08\%. Similarly, removing the domain discriminator loss leads to a reduction of 5.36\%. These results suggest that both discriminators contribute to effective feature decoupling. When both the domain and class discriminator losses are removed simultaneously, the performance decreases by 6.59\%, reflecting the complementary roles of the two discriminators in representation learning. Removing the Hierarchical-Domain Aggregation mechanism results in a performance drop of 4.47\%, highlighting its role in leveraging shared structures across subjects.}
\textcolor{black}{
The ablation results are summarized in Table~\ref{tab:Ablation}. Removing the domain prototype reduces the model to a conventional prototype learning framework, resulting in a performance drop of 5.75\%, which demonstrates the importance of subject-specific prototype modeling. Removing either the class or domain discriminator also leads to noticeable performance degradation, indicating that the two discriminators complement each other in effective feature disentanglement. When both discriminators are removed simultaneously, the performance decreases by 6.59\%, further confirming the necessity of disentangled representation learning. Moreover, removing the Hierarchical Domain Aggregation (HDA) mechanism results in a performance drop of 4.47\%, highlighting its effectiveness in capturing shared structures across subjects, which is further supported by the visualization results in Fig.~\ref{fig:Domain_feature}.
}
The removal of the adaptive update parameter $\alpha$ leads to a performance decrease of 3.16\%, suggesting that dynamic prototype updating contributes to stable model optimization. Replacing the pairwise learning strategy with conventional pointwise learning causes the largest performance degradation of 7.97\%, indicating the effectiveness of pairwise learning in improving robustness. 
\textcolor{black}{
In the Table~\ref{tab:Ablation}, replacing the learned bilinear matrix $\theta$ with an identity matrix decreases the accuracy from 84.70\% to 82.94\%, an absolute reduction of 1.76 percentage points. This result demonstrates that learning a task-specific compatibility function provides more effective sample-to-prototype matching than using a fixed inner-product similarity.}
%We remove $\theta$ and replace it with the identity matrix, which resulted in a performance decrease of 1.76\%. The results show that compared with the identity matrix, the trainable transformation matrix is more beneficial to enhance the feature representation ability of downstream tasks.} 
%我们删除$\theta$，并将其替换为单位矩阵，导致性能下降1.76\%，结果表明，与单位矩阵相比，可训练的转换矩阵更有利于增强下游任务的特征表示能力。。
Finally, removing the soft regularization term $\mathcal{R}$ leads to a moderate performance reduction, indicating its role in preventing redundant feature learning. Overall, the ablation results indicate that each component of the MAT framework contributes to the final performance, and their combination leads to more effective emotion recognition.

\begin{figure*}[t]
\centering
\includegraphics[width=1\textwidth]{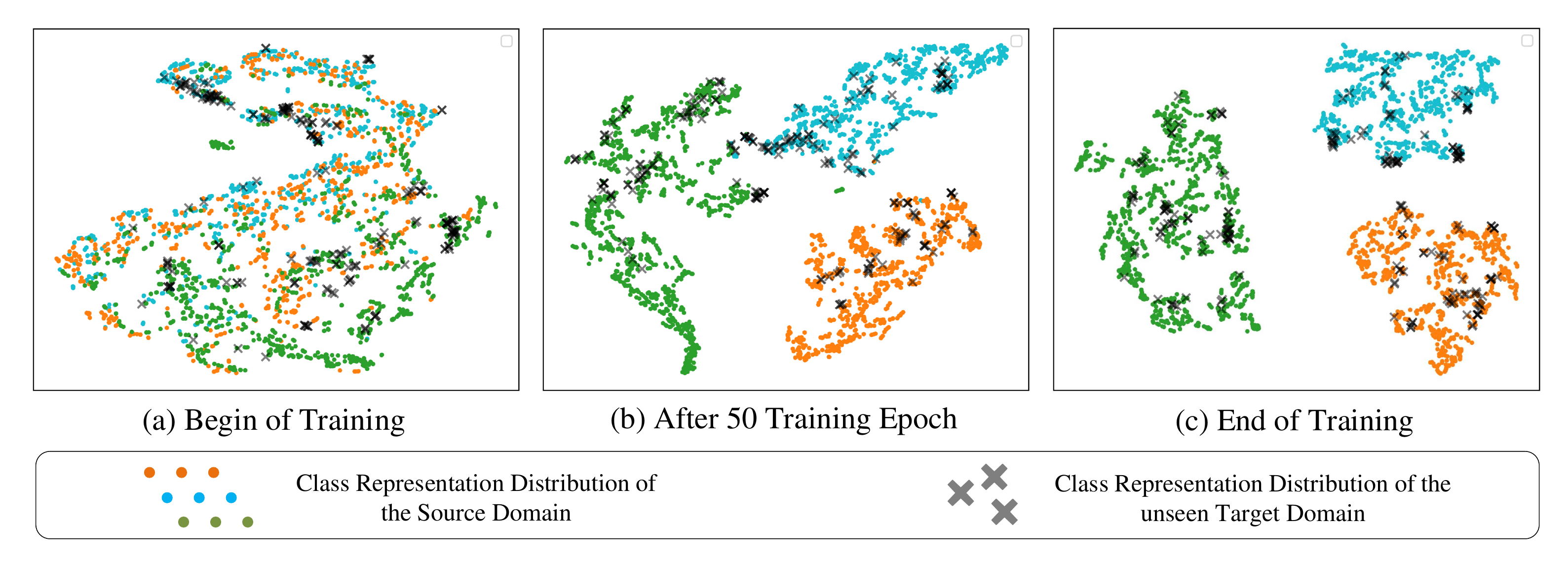}
\caption{T-SNE visualizations of class representation and class alignment of the MAT model, showing the distribution of class features at the beginning of training, after 50 training epoch, and end of training, respectively.}
\label{fig:Class_feature}
\end{figure*}
%----------------------------------------------------------------------

\subsection{Visualization of Domain and Class Representation}

\textcolor{black}{\indent To provide an intuitive analysis of the domain and class representations learned by the proposed MAT model and the hierarchical domain aggregation (HDA) mechanism, we visualize the feature distributions using t-SNE on the SEED dataset, illustrating the evolution of representations and prototypes during training.}

\textcolor{black}{As shown in Fig.~\ref{fig:Domain_feature}, the domain-specific features and corresponding domain prototype representations learned by MAT are visualized. Different colors denote features from different domains, the symbol $\diamond$ indicates the domain prototype of each domain, and the black $\times$ denotes features from the unseen target domain. Figures (a) to (c) present the results obtained without the HDA mechanism. In this setting, the number of domain clusters is relatively large, and the feature distributions exhibit substantial overlap, indicating that inter-domain relationships are not fully exploited. As a result, the utilization of source-domain information is limited. Figures (d) to (f) illustrate the results obtained with the HDA mechanism. By aggregating related domains into superdomains, the learned domain features become more compact and form clearer boundaries in the embedding space. 
\textcolor{black}{
In addition, variations in cluster sizes across superdomains can be observed, reflecting differences in shared feature distributions among subjects. Overall, the disentangled domain representations determine prototype construction, while HDA further regularizes prototype organization. The HDA mechanism effectively alleviates prototype overfitting and enhances the robustness of the learned representations across subjects.}  
%此外，可以观察到跨超域的簇大小变化，反映了受试者之间共享特征分布的差异。总的来说，分解后的域表示决定了原型的构造，HDA进一步规范了原型的组织。HDA机制有效解决了原型过拟合问题，增强了性能鲁棒性。
Furthermore, Fig.~\ref{fig:Class_feature} presents the visualization of class-specific feature distributions. As training progresses, samples belonging to the same emotion class gradually become more concentrated, while features from different classes are increasingly separated. This trend indicates improved class separability in the learned representations.
}

%----------------------------------------------------------------------
\begin{figure*}[h]
\centering
\subfloat{\includegraphics[width=1\textwidth]{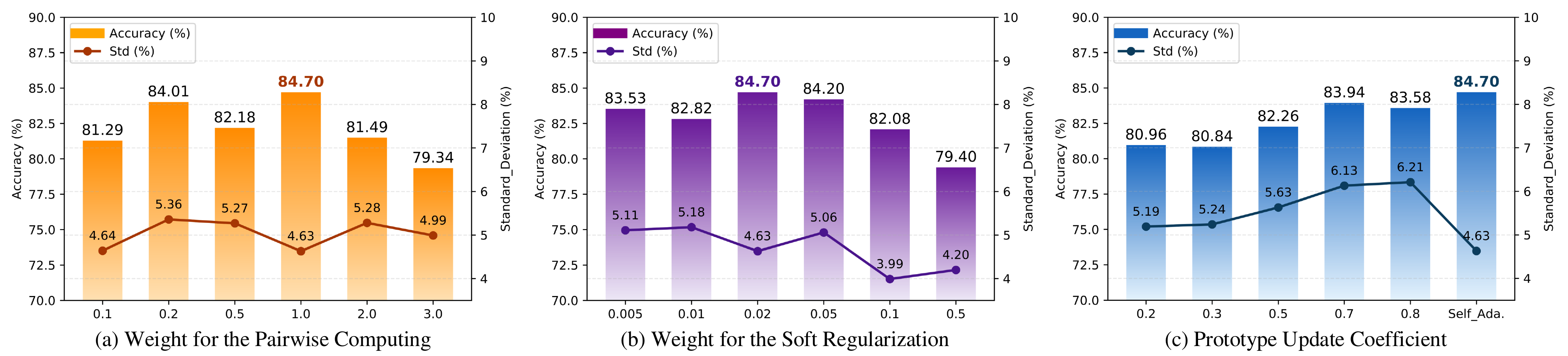}}
\caption{\textcolor{black}{Results for different parameter settings of the proposed MAT model.}}
\label{fig:Parameter Sensitivity}
\end{figure*}
%---------------------------------------------------------------------- 
%-----------------------------------------------------------------------
\begin{table}
\begin{center}
\caption{Results of the MAT model adding different proportions ($\eta\%$) of random label noise to the source domain, expressed as (Mean-Accuracy\% ± Standard-Deviation\%). \textcolor{orange}{$\uparrow$} denotes the performance differences.}
\label{tab:noisy labels}
\scalebox{1}{
\begin{tabular}{lccc}
\toprule
\textit{Noisy}  & \textit{Pointwise}  & \textit{Pairwise} & \textit{Pair. $-$ Point.}\\ 
\textit{Ratio} $(\eta)$  & \textit{Learning} (\%)  & \textit{Learning} (\%) & (\%)\\
\midrule
0\%             &  76.73 ± 06.62     &  84.70 ± 04.63  
                & \textcolor{orange}{$\uparrow$ 07.97} \\
5\%             &  75.89 ± 06.75     &  83.61 ± 05.28 
                & \textcolor{orange}{$\uparrow$ 07.72} \\
10\%            &  73.45 ± 05.94     &  83.02 ± 05.06 
                & \textcolor{orange}{$\uparrow$ 09.57} \\
20\%            &  71.84 ± 07.96     &  82.31 ± 04.94 
                & \textcolor{orange}{$\uparrow$ 10.47} \\
30\%            &  70.04 ± 07.63     &  81.38 ± 05.68 
                & \textcolor{orange}{$\uparrow$ 11.34} \\
\midrule
30\% $-$ 0\%  & \textcolor{orange}{$\downarrow$ 06.69}
                & \textcolor{orange}{$\downarrow$ 03.32} \\
\bottomrule
\end{tabular}
}
\end{center}
\end{table}
%--------------------------------------------------------------------------- 

\subsection{Effect of Noisy Labels}

\textcolor{black}{\indent To evaluate the robustness of the proposed model under noisy supervision, we investigate the impact of label noise on both pairwise learning and pointwise learning strategies. We introduce synthetic symmetric label noise into the source domain by randomly replacing a proportion \(\eta\%\) of the ground-truth labels with incorrect labels.  We consider noise ratios of $\eta = 5\%, 10\%, 20\%$, and $30\%$.}
%为了评估所提出模型在噪声监督下的鲁棒性，研究了标签噪声对成对学习和逐点学习策略的影响。通过用不正确的标签随机替换一部分\(\eta\%\)的真实标签，将合成的对称标签噪声引入源域。我们考虑$\eta = 5\%, 10\%, 20\%$和$30\%$的噪声比率。

\textcolor{black}{The results are reported in Table~\ref{tab:noisy labels}. Under the pointwise learning strategy, the model achieves accuracies of 76.73\%, 75.89\%, 73.45\%, 71.84\%, and 70.04\% for noise ratios of $\eta = 0\%, 5\%, 10\%, 20\%$, and $30\%$, respectively. When the noise ratio increases to 30\%, the performance decreases by 6.69\%, indicating a notable sensitivity of pointwise learning to label noise. In contrast, the pairwise learning strategy exhibits improved robustness. As the noise ratio increases from 0\% to 30\%, the performance decreases from 84.70\% to 81.38\%, corresponding to a total reduction of 3.32\%. \textcolor{black}{It is worth noting that this synthetic symmetrical noise may not fully capture the structural ambiguity caused by individual differences in emotional arousal, but all these results indicate that pairwise learning is more resilient to noise labels.}}
%%值得注意的是，这种合成的对称噪声可能不能完全捕捉到情绪激发的个体差异所产生的结构性歧义，但所有这些结果均表明了成对学习对噪声标签更具弹性。
\textcolor{black}{
Overall, the proposed Relative Consistency Computing strategy reformulates single-sample classification as similarity estimation between sample pairs, making MAT more robust to label noise and improving its stability in practical application scenarios.}
%总的来说，所提出相关一致性计算方法将单一样本识别转换为样本间的相似性问题，这使得MAT在标签噪声水平较高的情况下依旧保持稳定。

\subsection{Parameter Sensitivity}

\textcolor{black}{\indent We further quantitatively evaluate the sensitivity and effectiveness of the proposed MAT framework under different parameters. The total loss function for MAT (Eq.\ref{Eq:MAT loss}) has two important hyperparameters: the weight $\alpha$ of the pairwise learning loss and the weight $\beta$ of the soft regularization, which both affect the performance of the MAT. As shown in Fig.~\ref{fig:Parameter Sensitivity} (a)$\sim$(b), we conducted a strict evaluation and achieved the best performance when $\alpha$ = 1.0 and $\beta$ = 0.02. Additionally, to verify the significance of adaptive prototype update strategy (Eq.~\ref{Eq:mu_d^t}) of the MAT, we evaluate the performance impact of dynamic update coefficient $\alpha$. The results are shown in Fig.~\ref{fig:Parameter Sensitivity} (c), MAT takes the best performance when $\alpha$ is self\_adaptive updated. However, when $\alpha$ is set to a fixed value, the model performance declines, which further proves that adaptive prototype updates contribute to the stability of model.}

\begin{figure}[t]
\centering
\subfloat{\includegraphics[width=0.48\textwidth]{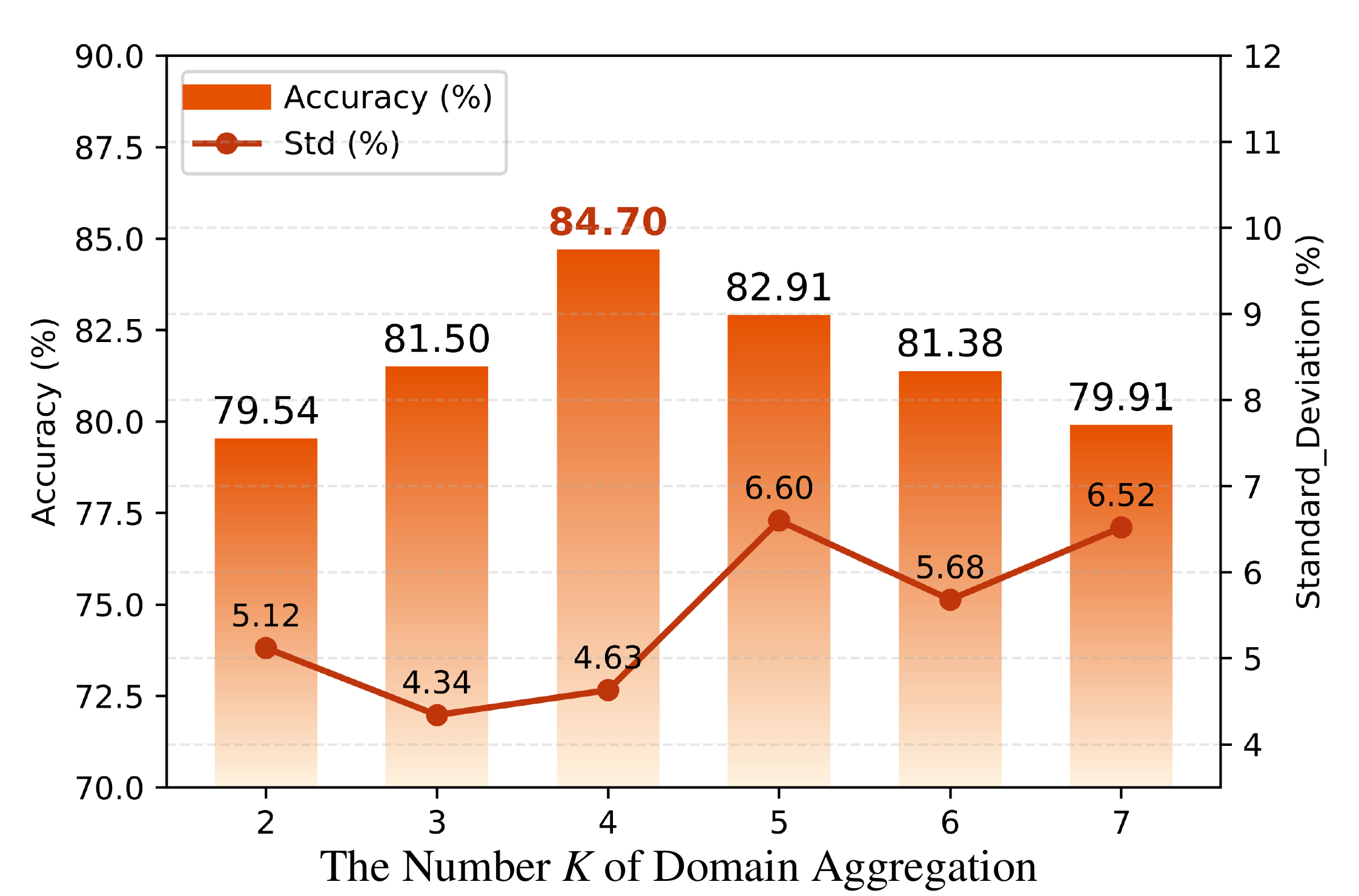}}
\caption{\textcolor{black}{Performance of the MAT model in different number of aggregates $K$.}}
\label{fig:Superdomain_number}
\end{figure}
%----------------------------------------------------------------------  

\subsection{Superdomain Number Analysis}
 
\textcolor{black}{\indent To analyze the influence of the superdomain number $K$ in the multi-domain aggregation mechanism, we examine the performance of MAT under different $K$ settings. \textcolor{black}{As shown in Fig.~\ref{fig:Superdomain_number}, the model achieves the highest accuracy of 84.70\% when $K=4$. Based on this observation, we set $K=4$ in all experiments. 
When $K$ is too small, source domains with low distributional similarity may be forced into the same superdomain, weakening the discriminability of the resulting prototypes. Conversely, an excessively large $K$ may split similar source domains across multiple superdomains, causing prototype fragmentation and less reliable routing of unseen samples. These results indicate that an appropriate superdomain granularity is important for reliable routing and subsequent class-prototype inference.} 
This phenomenon is consistent with the feature visualizations shown in Fig.~\ref{fig:Domain_feature} (c) and (f).}  

%------------------------------------------------------------------------------
\begin{table}[t]%MDD
\color{black}
\begin{center}
\caption{\textcolor{black}{Cross-subject LOSO cross-validation results on MDD dataset, expressed as (Acc\% ± Std\%). Here, '*' indicates the results are obtained by our own implementation. '\#' denoted as the model of the unseen target domain and  \textcolor{orange}{$\uparrow$} represents the gap between MAT and them.}}
\label{tab:MDD}
\scalebox{1}{
\begin{tabular}{lc|lc}
\toprule
Methods                     & $P_{acc}(\%)$ & 
Methods                     & $P_{acc}(\%)$ \\
\midrule
\multicolumn{4}{c}{\textit{Traditional machine learning methods}
                    \textcolor{orange}{($\uparrow$ 9.60\%)}} \\ 
\midrule
KNN*\cite{KNN1982}          & 70.22 ± 5.95 & 
KPCA*\cite{KPCA1999}        & \underline{70.56} ± 3.27 \\
SVM*\cite{SVM}              & 68.31 ± 7.98 & 
SA*\cite{SA2013}            & 69.31 ± 6.36 \\
TCA*\cite{TCA2010}          & 67.25 ± 5.38 & 
CORAL*\cite{CORAL2016}      & 67.47 ± 5.70 \\
GFK*\cite{GFK2012}          & 66.31 ± 5.39 & 
RF*\cite{breiman2001random} & 67.88 ± 4.67 \\
\midrule
\multicolumn{4}{c}{\textit{Deep transfer learning methods}
                    \textcolor{orange}{($\uparrow$ 4.99\%)}}\\ 
\midrule
DAN*\cite{He2018DAN}             & 76.28 ± 3.93 & 
DANN*\cite{ganin2016DANN}        & 74.07 ± 2.72 \\
DCORAL*\cite{Dcoral2016}         & 70.12 ± 3.71 & 
PL-DCP\#*\cite{LI2026DG}         & 71.76 ± 4.75 \\
EEG-DG\#*\cite{10609514_2024}    & \underline{75.17} ± 4.03 & 
DMMR\#*\cite{wang2024dmmr}       & 75.12 ± 4.52 \\
\midrule
\multicolumn{3}{l}{\textbf{MAT}} & \textbf{80.16 + 2.37}\\
\bottomrule
\end{tabular}
}
\end{center}
\end{table}
%------------------------------------------------------------------------------

\subsection{\textcolor{black}{Practical Applications}}
\indent
\indent \textcolor{black}{
Major depressive disorder (MDD) is closely associated with dysfunction of the emotion regulation system. To further evaluate the practical applications of MAT, we conducted experiments on an EEG-based clinical MDD dataset collected from HUSM \cite{mumtaz2015review}. After quality control, the dataset included 21 healthy controls and 31 patients with MDD. \textcolor{black}{The details are provided in Appendix G. As shown in Table~\ref{tab:MDD}, MAT achieve an accuracy of 80.16\%, outperforming the best deep-learning baseline (EEG-DG, 75.17\%) by 4.99\% and the best traditional machine-learning baseline (KPCA, 70.56\%) by 9.60\%. These results demonstrate the potential of the MAT framework for migration to practical applications.}}

%Major depressive disorder (MDD) is a serious mental illness. MDD is closely associated with dysfunction of the emotion regulation system, manifested as persistent negative affect, which constitutes one of the core diagnostic criteria for MDD. Therefore, we extend the MAT model to the actual scenarios of mental illness detection. We adopted an EEG-based clinical MDD database \cite{mumtaz2015review}, which was obtained from the Hospital University Sains Malaysia (HUSM). After careful screening to remove some defective data, our experiment included 21 healthy subjects and 31 subjects diagnosed with MDD. The experimental results are shown in Fig.~\ref{tab:MDD}, and all results are presented as the average accuracy of 10-fold cross validation. MAT achieved a performance of 80.16\%. Compared with the suboptimal result of deep-learning methods (EEG-DG, 75.17\%), the performance improved by 4.99\%. Compared with the suboptimal results of machine-learning methods (KPCA, 70.56\%), the performance of MAT has improved by 9.6\%. These results demonstrate the application potential of MAT in real clinical testing tasks.

\section{Conclusion}
%\textcolor{black}{\indent This work presents a multi-domain aggregation transfer learning framework with domain-class prototypes, referred to as MAT, for EEG-based emotion recognition under unseen target conditions. The proposed framework alleviates the reliance of conventional transfer learning approaches on target-domain data by disentangling domain-specific and class-specific representations. Specifically, a feature decoupling module is introduced to separate domain-related and emotion-related characteristics, while a hierarchical-domain aggregation mechanism is employed to adaptively learn domain and class prototype representations across subjects. In addition, a pairwise learning strategy is incorporated to reduce the sensitivity of the model to noisy emotion labels. Experimental results on multiple public EEG emotion datasets demonstrate that MAT achieves competitive performance compared with existing methods, despite the absence of target-domain data during training. These findings suggest that the proposed framework provides an effective solution for cross-subject EEG emotion recognition in practical scenarios where target-domain data are unavailable.}

\textcolor{black}{
\indent This work proposes a Multi-domain Aggregation Transfer learning framework with domain-class prototypes (MAT) for EEG emotion recognition under completely unseen target domains. By integrating feature disentanglement, hierarchical domain aggregation, adaptive prototype learning, and relative consistency learning, MAT effectively improves domain generalization while demonstrating strong robustness under challenging conditions. Moreover, the proposed framework generalizes well to clinical mental illness detection, highlighting its practical potential for real-world EEG analysis where target-domain data are unavailable.}
%This work proposed a Multi-domain Aggregation Transfer learning framework with domain–class prototypes (MAT) for emotion recognition under completely unseen target domains. The framework decouples domain-specific and class-specific representations, and the decomposed representations determine the construction of prototypes. The HDA mechanism further regularizes the organization of prototypes. MAT effectively enhances the domain generalization performance of traditional transfer learning methods, and is successfully extended to clinical practical applications. This framework provides an effective solution for emotion recognition in actual scenarios where target domain data is unavailable.}
%这项工作提出了一个具有领域类原型的多领域聚合迁移学习框架MAT，用于在未知目标条件下基于脑电图的情感识别。该框架分离出特定领域和特定类别的表示，分解后的表示决定了原型的构造，层次-领域聚合机制进一步规范了原型的组织。MAT有效增强了传统迁移学习方法的域泛化性能，并成功推广到了临床实际应用。该框架为目标域数据不可用的实际场景下的跨主体EEG情绪识别提供了有效的解决方案。

\section*{Acknowledgements}
\textcolor{black}{\indent This work was supported in part by the National Natural Science Foundation of China under Grant 62176089, 62522608, and 62276169, in part by the Key Research and Development Program of Hunan Province (2025QK3008), in part by the Guangdong Provincial Project (Grant No. 2024TQ08X330), in part by the Natural Science Foundation of Hunan Province under Grant 2023JJ20024, in part by the Key Research and Development Project of Hunan Province under Grant 2025QK3008, in part by the Key Project of Xiangjiang Laboratory under Granted 23XJ02006, in part by the Shenzhen Science and Technology Program (No. JCYJ20241202124222027 and JCYJ20241202124209011), and in part by Shenzhen-Hong Kong Institute of Brain Science-Shenzhen Fundamental Research Institutions (2023SHIBS0003).}

% or


%\appendices

%\section{}

% % use section* for acknowledgment
% \ifCLASSOPTIONcompsoc
%   % The Computer Society usually uses the plural form
%   \section*{Acknowledgments}
% \else
%   % regular IEEE prefers the singular form
% \fi

\ifCLASSOPTIONcaptionsoff
  \newpage
\fi

\bibliographystyle{IEEEtran}
\bibliography{references}

\begin{IEEEbiography}[{\includegraphics[width=1in,height=1.25in,clip,keepaspectratio]{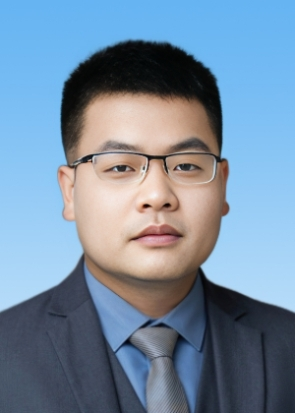}}]{Guangli Li}
\textcolor{black}{
is an Associate Professor and Doctoral Supervisor at the Hunan Key Laboratory of Biomedical Nanomaterials and Devices, School of Biological Science and Medical Engineering, Hunan University of Technology. He earned his Ph.D. in Physical Chemistry from Wuhan University in 2016. Notably, Dr. Li has been recognized as one of the world's top 2\% scientists since 2023. His research primarily focuses on ultrasensitive bioelectrochemical sensors, wearable flexible electronics, and non-invasive brain-computer interfaces.
 }
\vspace{-10mm}
\end{IEEEbiography}

\begin{IEEEbiography}[{\includegraphics[width=1in,height=1.25in,clip,keepaspectratio]{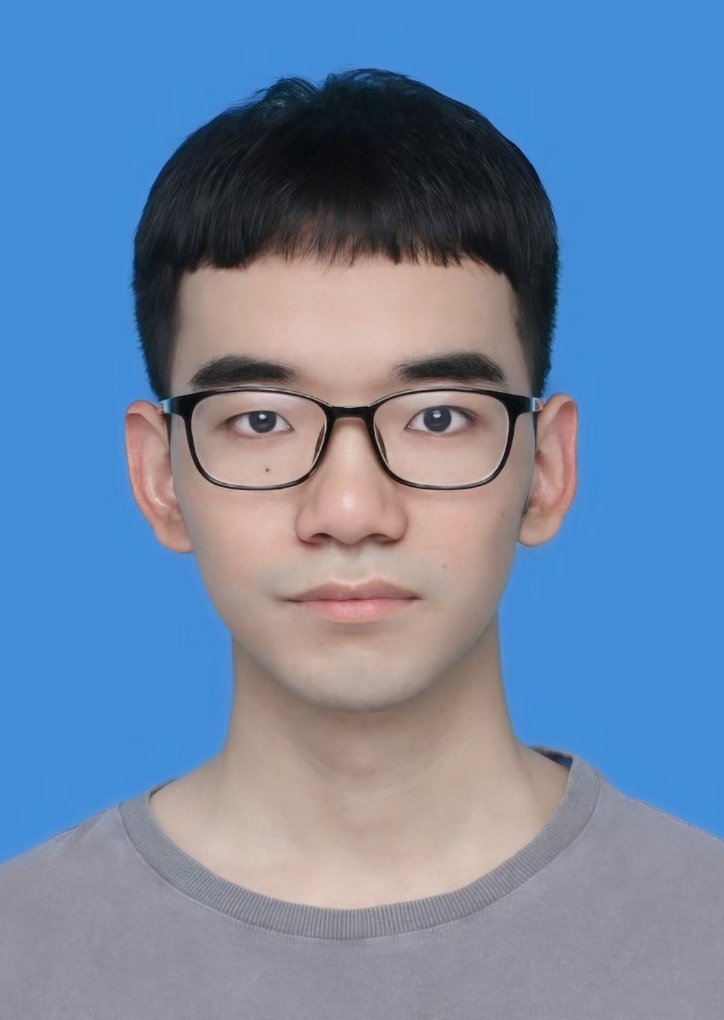}}]{Canbiao Wu}
received his bachelor’s degree in computer science and technology from the South China Normal University in 2022, and worked at the Institute for Brain Research and Rehabilitation in 2022 to 2023. He is currently a master’s student in Biomedical Engineering at Hunan University of Technology. His current research interests include affective BCI and EEG Large Model.
% \vspace{-16mm}
\vspace{-10mm}
\end{IEEEbiography}

\begin{IEEEbiography}[{\includegraphics[width=1in,height=1.25in,clip,keepaspectratio]{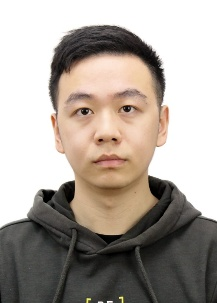}}]{Zhehao Zhou}
received his Master's degree in Biomedical Engineering from Hunan University of Technology in 2025. His research focused on metric learning and prototype learning.
\vspace{-10mm}
\end{IEEEbiography}

\begin{IEEEbiography}[{\includegraphics[width=1in,height=1.25in,clip,keepaspectratio]{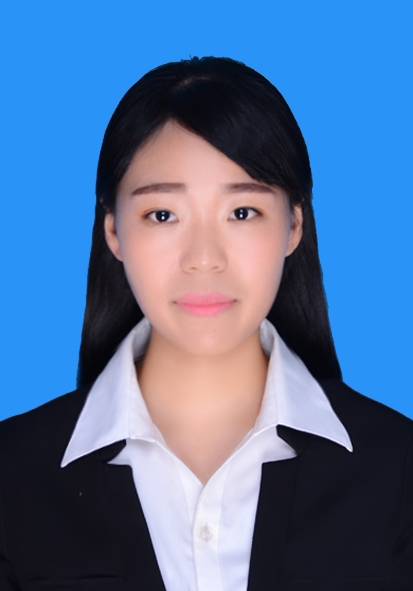}}]{Na Tian}
received M.Eng. and Ph.D. degrees in biomedical engineering from the Sun Yat-sen University in 2020 and 2024, respectively. She is currently a lecturer at the  School of Biological Science and Medical Engineering, Hunan University of Technology. Her research interests include biomedical signal processing, neuroregulation and rehabilitation engineering.
\vspace{-10mm}
\end{IEEEbiography}

\begin{IEEEbiography}[{\includegraphics[width=1in,height=1.25in,clip,keepaspectratio]{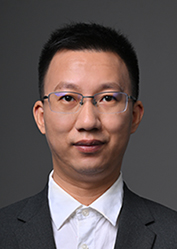}}]{Pengwen Dai}
received the Ph.D. degree from the Institute of Information Engineering, Chinese Academy of Sciences, in 2022. He is currently an associate professor with the School of Cyber Science and Technology, Sun Yat-sen University. His current research interests include multi-modal learning, computer vision, and pattern recognition.
\vspace{-10mm}
\end{IEEEbiography}

\begin{IEEEbiography}[{\includegraphics[width=1in,height=1.25in,clip,keepaspectratio]{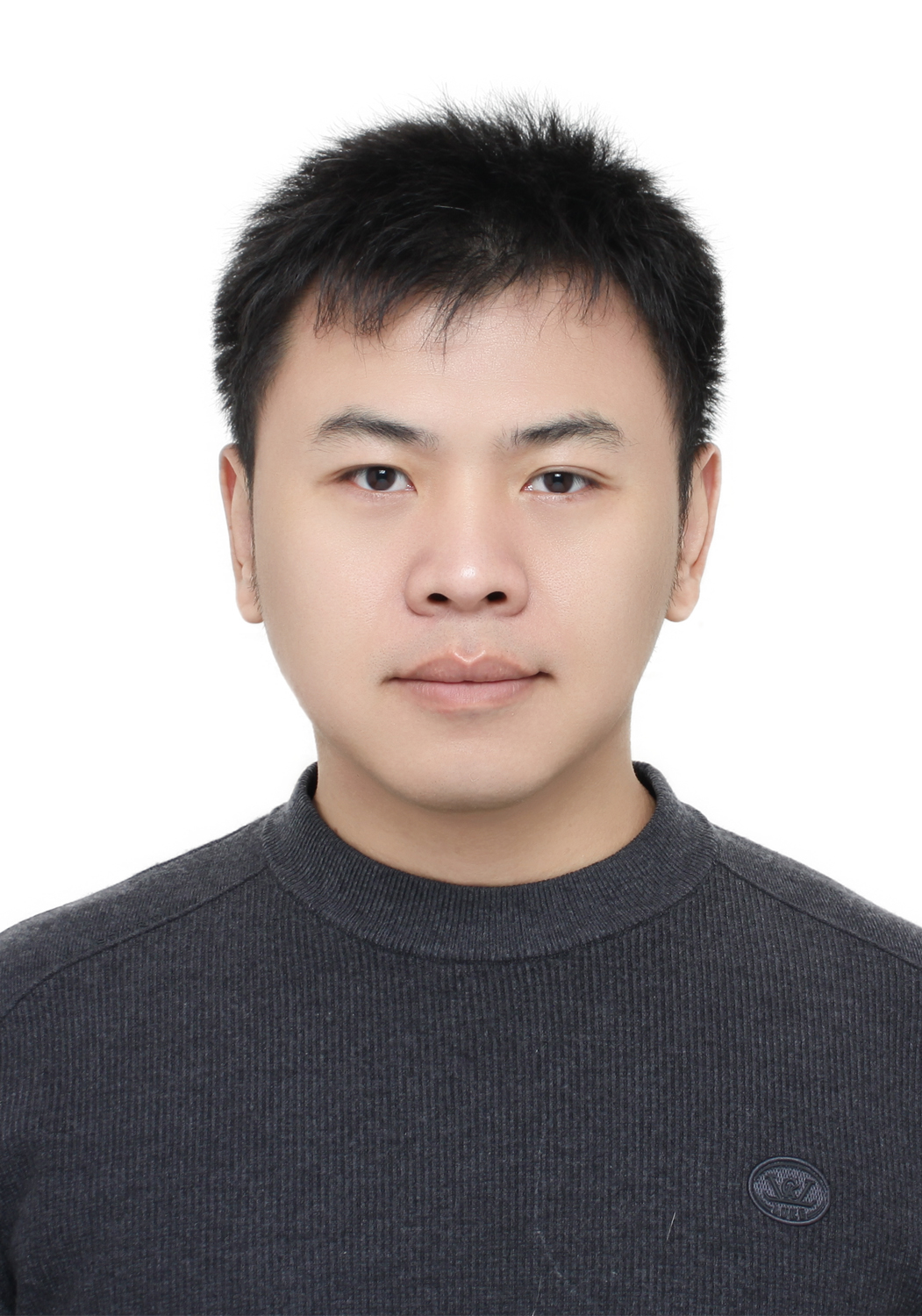}}]{Yuhao Wang}
received his Ph.D. degree in Medicine from The University of Osaka, Japan, in 2025. He is currently a postdoctoral researcher in the School of Computer Science and Technology, Beijing Institute of Technology, Beijing, China. His research interests include brain–computer interfaces, neurorehabilitation, and brain-inspired artificial intelligence.
\vspace{-10mm}
\end{IEEEbiography}

\begin{IEEEbiography}[{\includegraphics[width=1in,height=1.25in,clip,keepaspectratio]{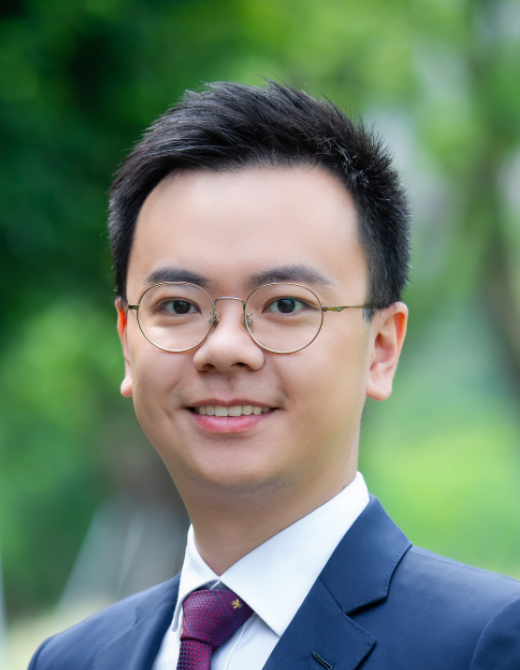}}]{Li Zhang}
received his B.Sc. degree in electronic information technology from the Macau University of Science and Technology in 2010, and the M.Sc. and Ph.D. degrees in electrical and electronic engineering from the University of Hong Kong in 2011 and 2017, respectively. He is currently an assistant professor at the School of Biomedical Engineering, Shenzhen University. His research interests include biomedical signal processing, brain imaging genetics and numerical optimization.
\vspace{-10mm}
\end{IEEEbiography}

\begin{IEEEbiography}[{\includegraphics[width=1in,height=1.25in,clip,keepaspectratio]{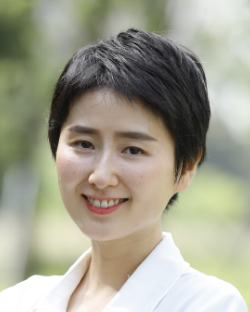}}]{Zhen Liang}
received her Ph.D. degree from The Hong Kong Polytechnic University, Hong Kong, in 2013. From 2012 to 2017, she was an algorithm development scientist at NeuroSky, Inc., Hong Kong. From 2018 to 2019, she was a specially‐appointed assistant professor of Graduate School of Informatics, Kyoto University, Japan. She is currently a distinguished professor in the School of Biomedical Engineering, Health Science Center, Shenzhen University, China. Her current research interests include brain-computer interface, brain encoding and decoding systems, affective computing, and neural engineering.
\vspace{-10mm}
\end{IEEEbiography}

\clearpage
{\Large \textbf{~~~Supplementary Materials}}

\subsection*{Appendix A: Hierarchical Domain Aggregation}

\textcolor{black}{\quad Unlike conventional domain adaptation methods that perform global distribution alignment across all source domains [S1][S2], the proposed Hierarchical Domain Aggregation (HDA) explicitly models the hierarchical relationships among subjects by organizing similar domains into superdomains. Prototype learning and adaptive prototype updates are then performed within each superdomain, enabling shared structure learning while preserving subject-specific characteristics. As shown in Fig.~S1 (a), if the HDA mechanism is not adopted, the decomposed domain features can only perform global distribution matching, which does not consider the hierarchical similarity between individual domains. This leads to additional computational cost and overfitting as the number of subjects increases [S3][S4]. As shown in Fig.~S1 (b), HDA mechanism provides subspace-aware regularization for prototype organizations. The prototype update process is restricted to each superdomain, domain prototypes aggregate shared individual patterns within the superdomain, while class prototypes are refined under a consistent superdomain space. In addition, we design the prototype update rules with dynamic adaptation to match this superdomain hierarchy.}

%-----------------------------------------------------------------------
\begin{figure*}[h]
\captionsetup{labelformat=empty}
\centering
\includegraphics[width=0.85\textwidth]{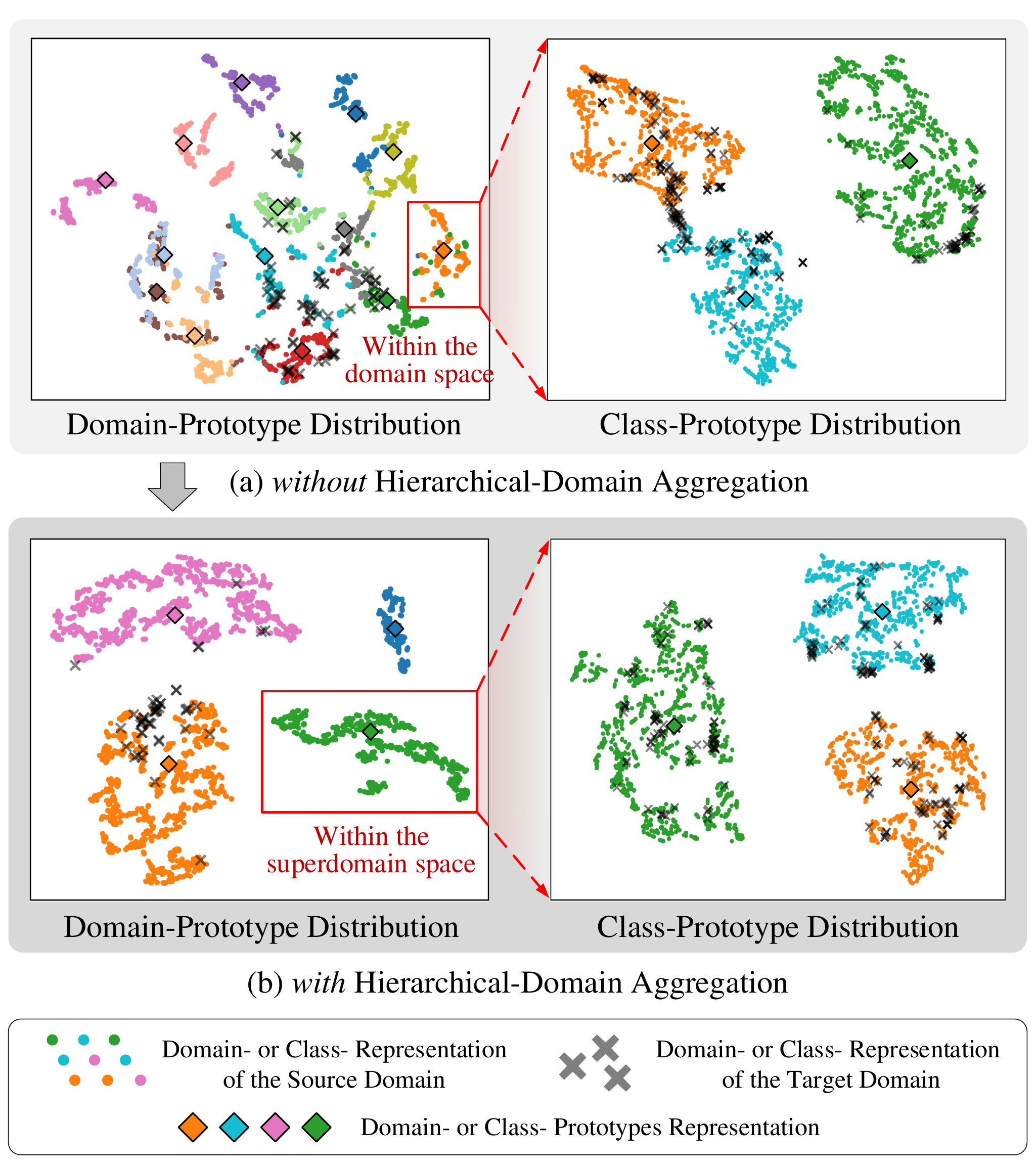}
\caption{\textcolor{black}{Figure~S1: The T-SNE visualization of the \textit{without} HDA and \textit{with} HDA.}}
\label{fig:HDA}
\end{figure*}
%-----------------------------------------------------------------------

\subsection*{Appendix B: Module Interaction and Joint Optimization}
 
\textcolor{black}{\quad To clear demonstrate the interaction among feature disentanglement, prototype learning, and hierarchical domain aggregation, as shown in Fig.S2, we provide a schematic illustration of model optimization. Specifically, these three modules are connected through top-down prior constraints and bottom-up iterative optimization. Feature disentanglement provides the representation subspaces for dual-prototype learning. Domain-specific and class-specific representations define the modeling boundaries of the corresponding prototypes, while prototype learning, in turn, supervises and refines the disentangled representations through iterative optimization. HDA mechanism further regularizes prototype learning by organizing similar subjects into superdomains. Instead of maintaining independent prototypes for each subject, prototype updates are performed within each superdomain, enabling shared structure learning while preserving subject-specific characteristics. Furthermore, iterative prototype optimization continuously refines the superdomain organization through MMD-based [S5] similarity estimation. Overall, feature disentanglement defines the representation space for prototype learning, prototype optimization further refines the disentangled representations, and HDA regularizes prototype organization by capturing shared structures across subjects. Together, these three modules form a unified optimization framework for robust cross-subject representation learning.}

%-----------------------------------------------------------------------
%\vspace{0em}
\begin{figure*}[h]
\captionsetup{labelformat=empty}
\centering
\includegraphics[width=1\textwidth]{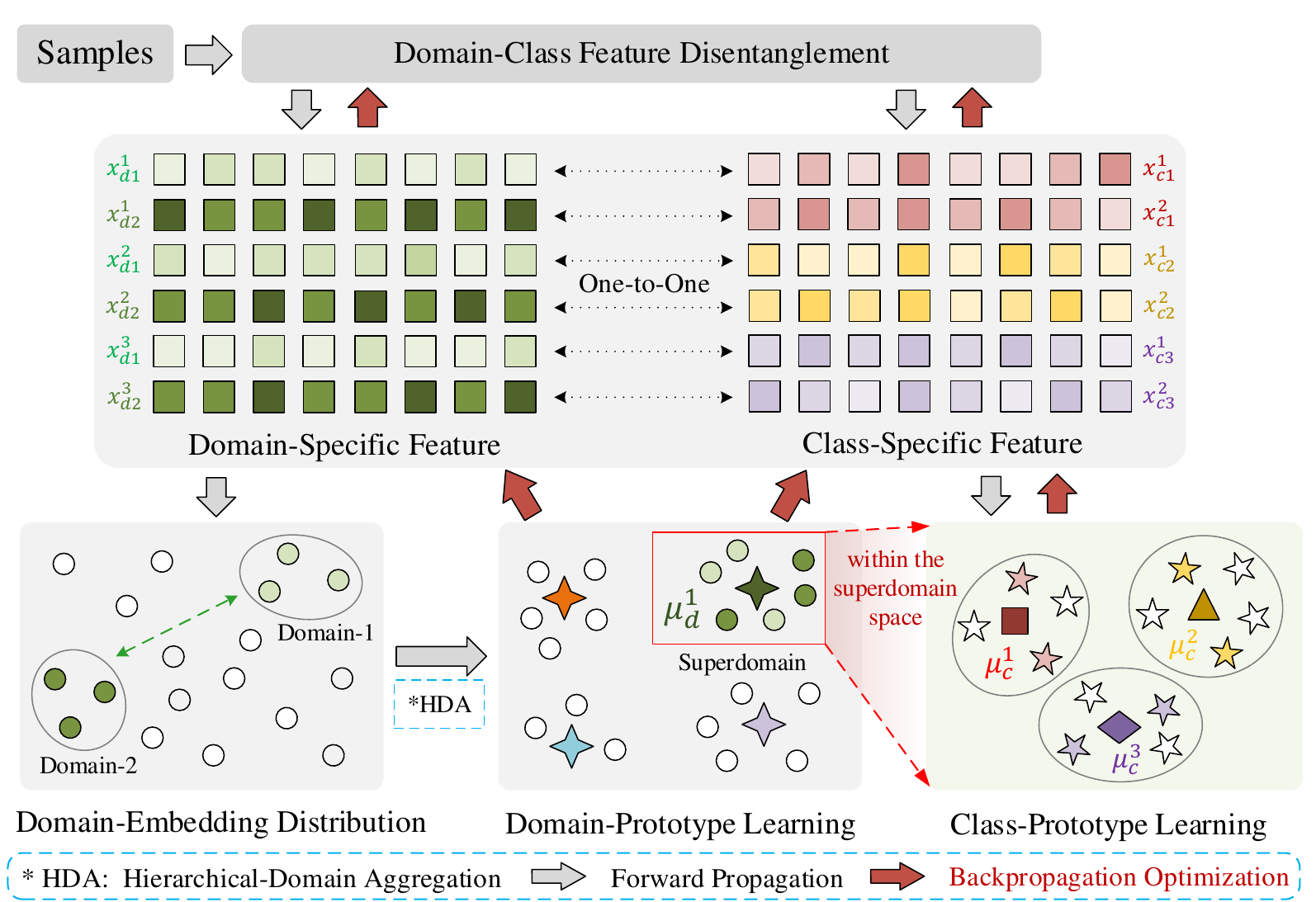}
\caption{\textcolor{black}{Figure~S2: Schematic illustration of the interaction and iterative optimization among domain-class disentanglement, prototype learning, and hierarchical domain aggregation. Each sample is decomposed into domain-specific and class-specific representations, which are in one-to-one correspondence. The illustrated samples originate from two different subjects and therefore exhibit distinct subject-specific characteristics (light green and dark green), while samples belonging to the same emotion share similar emotion-specific patterns. Accordingly, the three emotion categories from different subjects are represented by red, yellow, and purple, respectively.}}
\label{fig:Interaction}
\end{figure*}
%-----------------------------------------------------------------------

\subsection*{Appendix C: Loss Optimization Curve}

\quad To provide deeper insight into the optimization trend and convergence characteristics of the proposed MAT framework, we provide the loss curves for each key loss term during model training. As shown in Fig.~S3 (a)$\sim$(c), the disentanglement loss continues to converge with the training of the model and eventually remains stable, which indicates that the MAT model realizes the gradual separation of domain-class features, and the disentanglement process has excellent stability. The MAT model transforms the traditional classification problem into the sample-based similarity calculation problem. In Fig.~S3 (d), the continuous decrease of the loss curve indicates that the model is able to gradually learn a more discriminative similarity measure, thereby enhancing the recognition performance of the model. The total loss curve of MAT model is shown in Fig.~S3 (e). The overall trend decreases continuously and remains stable eventually, which further proves the optimization stability and convergence of proposed MAT model.

%----------------------------------------------------------------------
\begin{figure*}[h]
\captionsetup{labelformat=empty}
\centering
\subfloat{\includegraphics[width=1\textwidth]{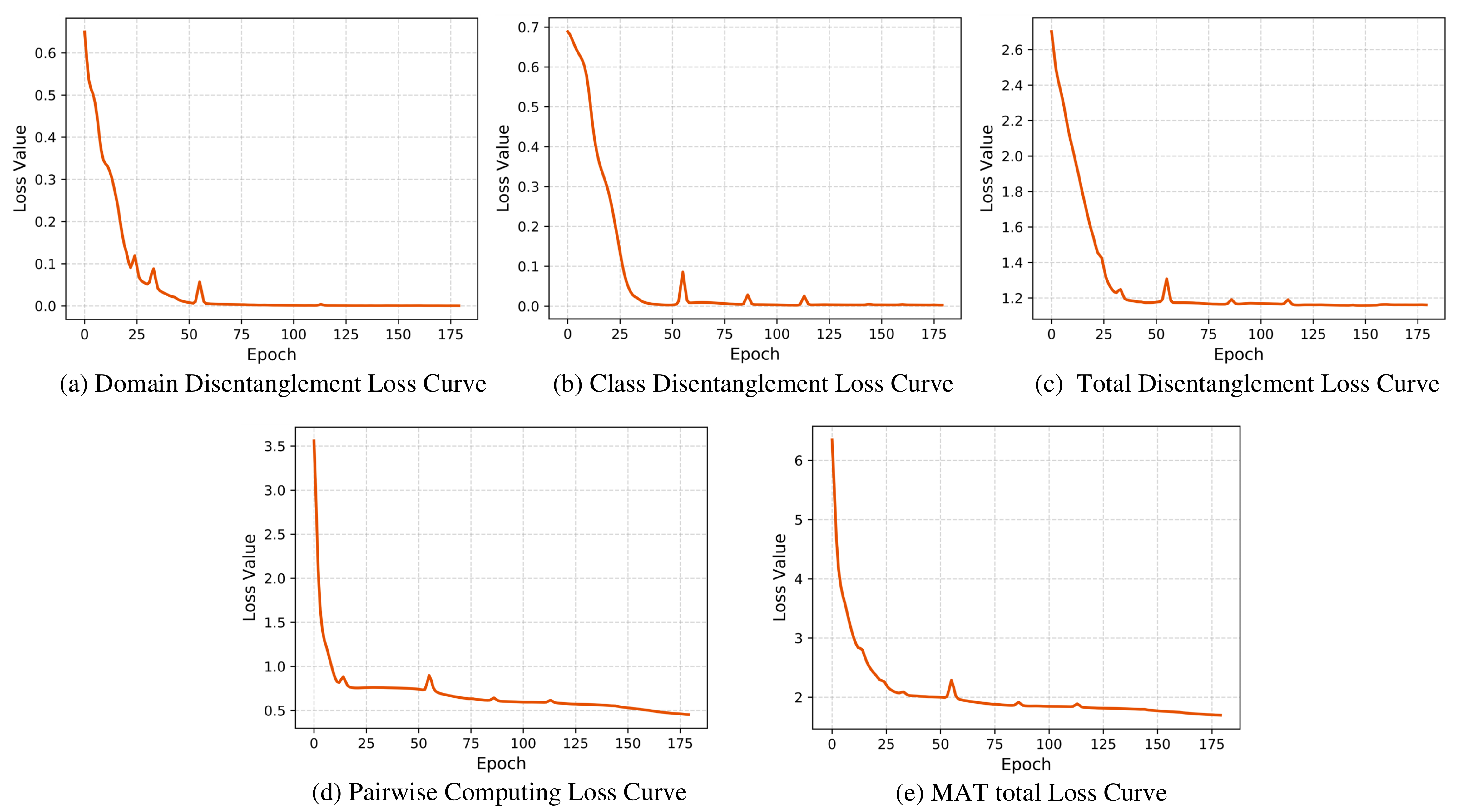}}
\caption{\textcolor{black}{Figure~S3: The loss curve of proposed MAT: (a) Domain feature disentanglement loss \textit{Eq.~4}. (b) Class feature disentanglement loss \textit{Eq.~3}. (c) Total disentanglement loss \textit{Eq.~5}. (d) Pairwise computing loss \textit{Eq.~17}. (e) The MAT total loss \textit{Eq.~19}.}}
\label{fig:Loss Curve}
\end{figure*}
%---------------------------------------------------------------------- 

\subsection*{Appendix D: Model Implementation Pseudocode}

\textcolor{black}{\quad To improve the clarity and repeatability of the MAT framework, as well as the update sequence of the each components, we provide the pseudo-code of the MAT framework (Algorithm~S1). The complete process of MAT is divided into the training phase and the inference phase. During the training phase, the model is completely independent of the target domain data. The ultimate optimization goal of MAT is to obtain the optimal domain prototype representation and class prototype representation. In the inference phase, MAT utilizes the learned domain class prototype representation for emotion recognition. Specifically, the MAT model first conducts domain prototype inference to determine the superdomain space to which the EEG data belongs. Then, in this hyperdomain space, MAT performs class prototype inference to determine the emotion label to which it belongs. The domain class dual prototype system effectively solves the limitations of individual differences and feature offset.}

%-----------------------------------------------------------------UTL-EADP_algorithm
\begin{algorithm*}[h]
\captionsetup{labelformat=empty}
\caption{\textcolor{black}{Algorithm S1: The MAT implementation process.}}
\label{Code_MAT}
%\setstretch{1}
\color{black}
\textbf{Input:} 
EEG Emotion Data ($\mathbb{S},\mathbb{T}$); Initialization parameters: a shallow feature extractor ($f_g$); domain-class feature decoupler ($f_d\ /\ f_c$); domain-class discriminator ($D_d\ /\ D_c$); Gradient Reversal Layer ($R$); Epoch\_Size($N$).
\\ \textbf{Training Phase:}\quad \textcolor{gray}{\# Only input Source Domain Data}
\begin{algorithmic}[1]    
    \For{1 to $N$}:
        \State \textcolor{gray}{\# Extract domain-class features}
        \item[] \quad\ \  $(x_d, x_c) \gets (f_d, f_c, f_g)$ by Eq.2; 
        \State \textcolor{gray}{\# Decouple class-specific features}
        \item[] \quad\ \  Adversarial optimization $\mathcal{L}_{cls} \gets (D_c, R)$ by Eq.~3;
        \State \textcolor{gray}{\# Decouple domain-specific features}
        \item[] \quad\ \  Adversarial optimization $\mathcal{L}_{dom} \gets (D_d, R)$ by Eq.~4; 
        \State \textcolor{gray}{\# Perform HDA mechanism}
        \item[] \quad\ \ Reorganized structure $\mathbb{S} \gets [S_1, S_2, \ldots, S_K]$ by Eq.~8$\sim$10;
        \State Capture domain-class prototype $(\mu_d, \mu_c) \gets (x_d^i, x_c^i)$ by Eq.~11;
        \State \textcolor{gray}{\# Noise-Robust Optimization}
        \item[] \quad\ \ Pairwise computing $\mathcal{L}_{pair} \gets (\mathcal{G}, x_c^i, x_c^j)$ by Eq.~17;
        \State Overall optimization $\mathcal{L}_{MAT} \gets (\mathcal{L}_{cls}, \mathcal{L}_{dom}, \mathcal{L}_{pair})$ by Eq.~19;
    \EndFor 
\end{algorithmic}
Save the optimal domain-class prototype ($\mu_d, \mu_c$).\\
\textbf{Inference Phase:}\quad \textcolor{gray}{\# Input Target Domain Data}
\begin{algorithmic}[1]
    \State Load the domain-class prototype ($\mu_d, \mu_c$);
    \State Match the optimal domain prototype space $\mathcal{V}_i \gets (x_d^i, \mu_d)$ by Eq.~15;
    \State Match class prototype $\mathcal{P}_i \gets (x_c^i,\mu_c)$ by Eq.~16;
\end{algorithmic}
Output emotional state.
\end{algorithm*}
%--------------------------------------------------------------------------------------

\subsection*{Appendix E: Confusion Matrix}

\textcolor{black}{\quad To qualitatively evaluate the recognition performance of MAT for different emotion categories, we visualize the confusion matrices of MAT model and baseline model (DANN [S6], DAN [S7], DCORAL [S8]). As shown in Fig.~S4 (a) to (d), on the SEED dataset, the positive emotion recognition accuracy of MAT is 86.35\%. Similar to other models, it has the best performance in recognizing positive emotions. In addition, the recognition differences between different emotions of the baseline models (DANN, DAN and DCORAL) are 6.19\%, 8.77\% and 10.4\%, while that of MAT is only 4.03\%, which indicates that the MAT model has relatively excellent stability. The confusion matrix on the SEED-IV dataset is shown in Fig.~S4 (e) to (h). All the participating models performed poorly in recognizing the emotion of happy, while they all performed well in recognizing the other three emotions. When identifying neutral and sad emotions, compared to the sub-optimal models DANN (65.96\%) and DCORAL (59.56\%), the performance of the MAT model was improved by 4.36\% and 11.59\%, respectively. When identifying fear and happy emotions, the performance of MAT was improved by 14.36\% and 2.25\% respectively compared with the suboptimal model DAN. As shown in Fig.~S5 (a) to (d), on the SEED-V dataset, the proposed MAT model has weak performance in identifying happy emotions, and is easy to confuse happy emotions with neutral emotions. On the contrary, MAT has the best performance in recognizing fear, sadness and nausea with 60.86\%, 67.72\%, and 75.00\%, respectively. Compared with the sub-optimal model, the performance of MAT is improved by 4.79\%, 7.46\%, and 11.45\%, respectively.}

%------------------------------------------------------------------------------
\begin{figure*}[h]
\captionsetup{labelformat=empty}
\centering
\includegraphics[width=1\textwidth]{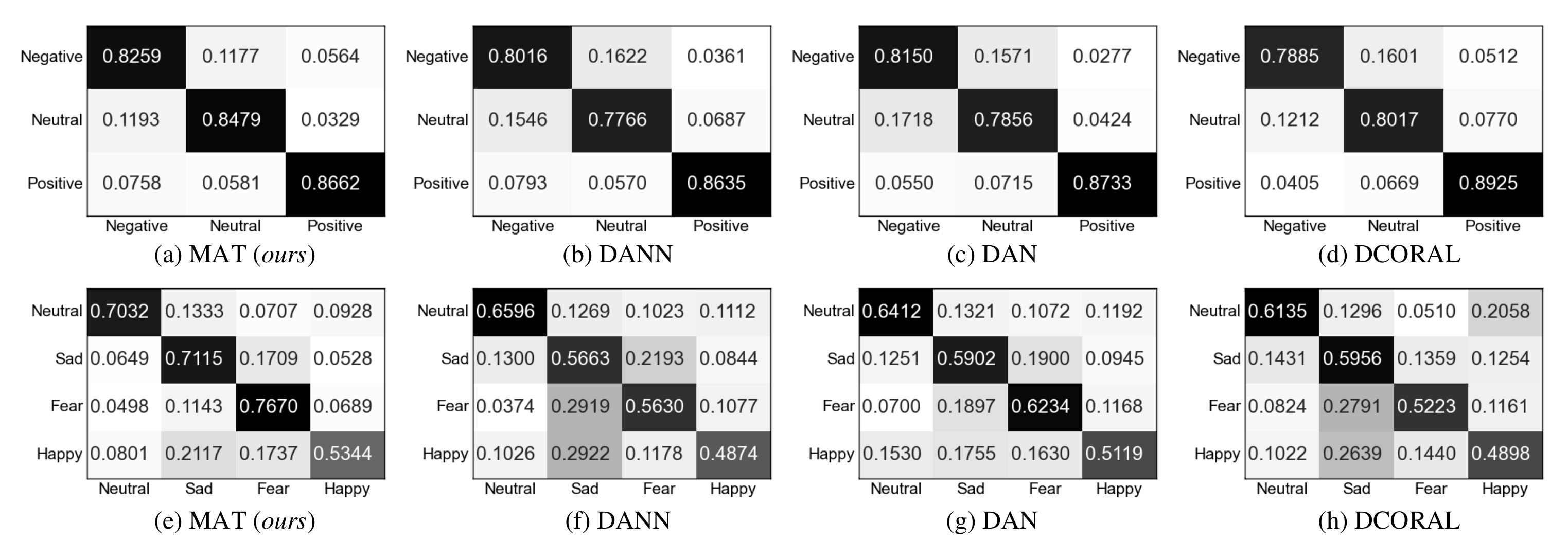}
\caption{\textcolor{black}{Figure~S4: Confusion matrices of different baseline models under cross-subject single-session LOSO cross-validation. The SEED database (a)$\sim$(d) contains three emotion categories: negative, neutral and positive. The Seed-IV database (e)$\sim$(h) contains four emotion categories: neutral, sad, fear and happy. The horizontal axis represents the predicted labels, while the vertical axis represents the true labels.}}
\label{fig:SEED_SEEDIV}
\end{figure*}
%------------------------------------------------------------------------------
\begin{figure*}[h]
\captionsetup{labelformat=empty}
\centering
\includegraphics[width=1\textwidth]{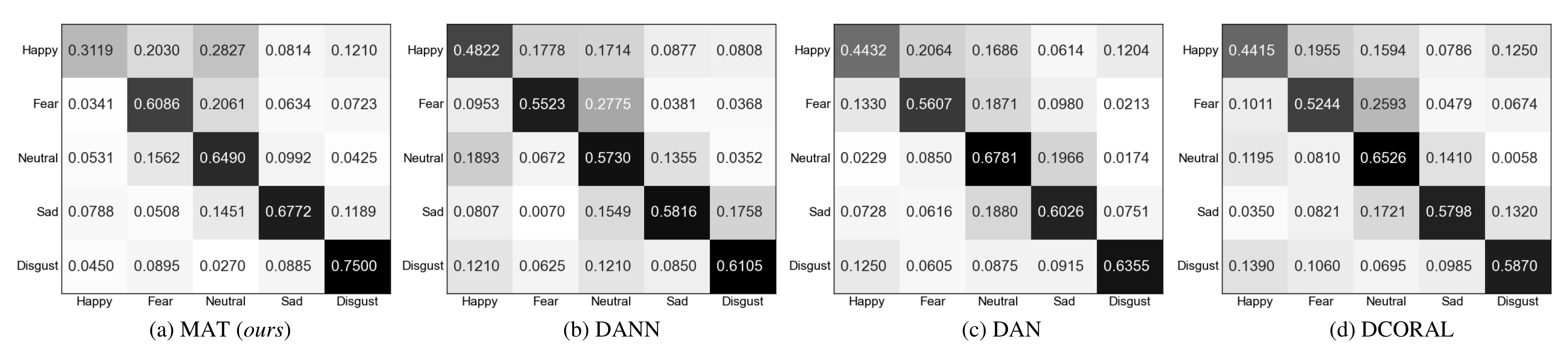}
\caption{\textcolor{black}{Figure~S5: Confusion matrices of different baseline models under cross-subject single-session LOSO cross-validation. The Seed-V database contains five emotion categories: happy, fear, neutral, sad, and disgust. The horizontal axis represents the predicted labels, while the vertical axis represents the true labels.}}
\label{fig:SEEDV}
\end{figure*}
%-----------------------------------------------------------------------------

\subsection*{Appendix F: Statistical Significance Test}

\textcolor{black}{\quad We provided a statistical significance test, which can confirm the effectiveness of the performance improvement of the MAT model. The results are shown in Fig.~S6. In all evaluation databases (SEED, SEED-IV, SEED-V, and SEED-VII), the MAT and machine learning baseline comparison results all achieved statistically significant performance improvements ($p<0.05$). In the SEED database, the results of the two sets of deep learning optimal baseline models are not implemented by us, and the p-values cannot be calculated because cross-validated performance results are not available. In addition, the performance improvement results (Single-Session: 44.69 ± 6.71 to 46.90 ± 7.72; Cross-Session: 31.67 ± 5.28 to 34.49 ± 4.22) of deep transfer learning baseline on SEED-VII data are marginally significant ($p<0.1$). In the remaining experimental configurations, the deep learning baseline comparison results show statistically significant performance improvement. Overall, these results verify the effectiveness of MAT on EEG emotion recognition tasks and provide a potential solution for emotion recognition under unseen target domain conditions.}

%-------------------------------------------
%\vspace{2em}
\begin{figure*}[h]
\captionsetup{labelformat=empty}
\centering
\includegraphics[width=0.96\textwidth]{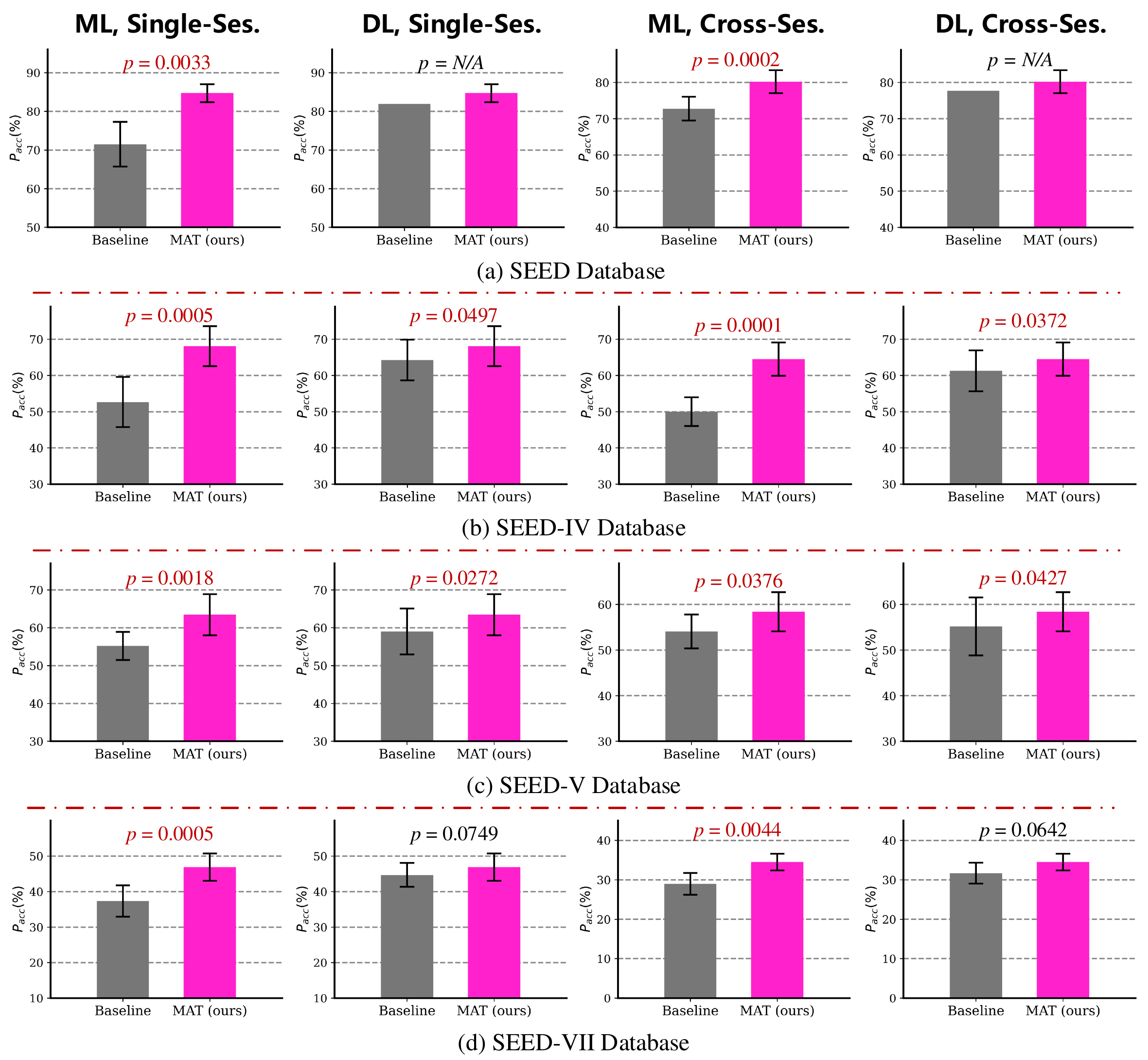}
\caption{\textcolor{black}{Figure S6: Obtained p-values of the paired t-test between the proposed MAT and the baseline method in terms of accuracy. The error bars represent the standard deviation of the obtained experimental results. Here, ML and DL represent the baseline comparison results of the machine learning model and the deep transfer learning model, respectively. N/A indicates that the p-value could not be obtained because the results of the baseline model were not implemented by us.}}
\end{figure*}
%-------------------------------------------

\subsection*{Appendix G: MDD Database Details}

\textcolor{black}{\quad The MDD database is an EEG-based clinical major depressive disorder database, which was obtained from the Hospital University Sains Malaysia (HUSM), which included 30 healthy subjects (38.28 ± 15.64 years old) and 34 subjects diagnosed with MDD (40.33 ± 12.86 years old) [S9]. Their MDD categorization adheres to the international diagnostic criteria for depression outlined in the Diagnostic and Statistical Manual-IV (DSM-IV). After quality control, the dataset included 21 healthy controls and 31 patients with MDD. The EEG data was collected using the BrainMaster Discovery amplifier, with 19 EEG electrodes anda sampling rate of 256Hz. (\textit{Data Sources: https://doi.org/10.1016/j.bspc.2015.07.003; Data download: https://doi.org/10.6084/m9.figshare.4244171)}}

In our work, the preprocessing process of MDD databases is similar to that of emotion databases (SEED, SEED-IV, SEED-V, and SEED-VII). We downsampled EEG signals to 200 Hz and manually removed interfering signals such as EMG and EOG. Subsequently, we used a band-pass filter (0.3-50 Hz) to filter the EEG signal and split it into 1-second non-overlapping segments. Based on the predefined frequency bands (\textit{Delta} 1-3 Hz, \textit{Theta} 4-7 Hz, \textit{Alpha} 8-13 Hz, \textit{Beta} 14-30 Hz and \textit{Gamma} 31-50 Hz), we extract the corresponding differential entropy (DE) features to represent the logarithm energy spectrum in a specific frequency band. Finally, the size of the preprocessed MDD data is (N, 19 channels $\times$ 5 bands). We divided healthy subjects and MDD subjects into source and target domains with a ratio of 7:3 without any data leakage.
\\[1em]
{\large \textbf{References}}\\[0.5em]
\textnormal{[S1]} R. Zhou, W. Ye, Z. Zhang, Y. Luo, L. Zhang, L. Li, G. Huang, Y. Dong, Y.-T. Zhang, and Z. Liang, “Eegmatch: Learning with incomplete labels for semisupervised eeg-based cross-subject emotion recognition,” IEEE Transactions on Neural Networks and Learning Systems, vol. 36, no. 7, pp. 12 991–13 005, 2025.\\[0.5em]
\textnormal{[S2]} W. Ye, Z. Zhang, F. Teng, M. Zhang, J. Wang, D. Ni, F. Li, P. Xu, and Z. Liang, “Semi-supervised dual-stream self-attentive adversarial graph contrastive learning for cross-subject eeg-based emotion recognition,” IEEE Transactions on Affective Computing, vol. 16,
no. 1, pp. 290–305, 2025.\\[0.5em]
\textnormal{[S3]} H. Zhao, S. Zhang, G. Wu, J. M. Moura, J. P. Costeira, and G. J. Gordon, “Adversarial multiple source domain adaptation,” Advances in neural information processing systems, vol. 31, 2018.\\[0.5em]
\textnormal{[S4]} M. Long, Z. Cao, J. Wang, and M. Jordan, “Conditional adversarial domain adaptation,” Advances in neural information processing systems, vol. 31, 2018.\\[0.5em]
\textnormal{[S5]} M. Arbel, A. Korba, A. Salim, and A. Gretton, “Maximum mean discrepancy gradient flow,” Advances in neural information processing systems, vol. 32, 2019.\\[0.5em]
\textnormal{[S6]} Y. Ganin, E. Ustinova, H. Ajakan, P. Germain, H. Larochelle, F. Laviolette, M. Marchand, and V. Lempitsky, “Domain-adversarial training of neural networks,” J. Mach. Learn. Res., vol. 17, no. 1, p. 2096–2030, 2016.\\[0.5em]
\textnormal{[S7]} H. Li, Y.-M. Jin, W.-L. Zheng, and B.-L. Lu, “Cross-subject emotion recognition using deep adaptation networks,” in International conference on neural information processing. Springer, 2018, pp. 403–413\\[0.5em]
\textnormal{[S8]} B. Sun and K. Saenko, “Deep coral: Correlation alignment for deep domain adaptation,” in European conference on computer vision (ECCV). Springer, 2016, pp. 443–450.\\[0.5em]
\textnormal{[S9]} Mumtaz W, Malik A S, Yasin M A M, et al. Review on EEG and ERP predictive biomarkers for major depressive disorder[J]. Biomedical Signal Processing and Control, 2015, 22: 85-98.

\end{document}